\documentclass{article}

\usepackage[preprint]{corl_2026} 

\usepackage{graphicx} 
\usepackage{amsmath,amssymb} 

\newcommand{\figref}[1]{Fig.~\ref{#1}}
\newcommand{\secref}[1]{Sec.~\ref{#1}}
\newcommand{\tabref}[1]{Tab.~\ref{#1}}
\renewcommand{\eqref}[1]{Eq.~\ref{#1}}

\title{TEMPO: Learning Temporal Context for Dynamic Robot Manipulation}
\author{
  Zhenyang Feng\thanks{Equal contribution.}\quad
  Jimin Heo\footnotemark[1]\quad
  Erik B.\ Sudderth\quad
  Unnat Jain \\
  University of California, Irvine \\
  \texttt{\{dfeng8, heoj4, sudderth, unnatj\}@uci.edu}
}

\usepackage{booktabs}
\usepackage[capitalize]{cleveref}
\usepackage{wrapfig}

\newcommand{\todo}[1]{\textbf{\color{red}[TODO: #1]}}

\definecolor{orange}{rgb}{0.85,0.45,0.0}
\definecolor{cyan}{rgb}{0.0,0.8,0.8}
\newcommand{\llm}[1]{{\color{orange}#1}}

\newcommand{\methodname}{Temporal Encoding for Motion-aware Policy (TEMPO)}
\newcommand{\ours}{TEMPO}
\newcommand{\TEMPOMOT}{TEMPO\textsubscript{MOT}}
\newcommand{\TEMPOACT}{TEMPO\textsubscript{ACT}}
\newcommand{\benchname}{TEMPO-Bench}

\begin{document}
\maketitle
\vspace{-2em}


\begin{abstract}

Vision-language-action (VLA) models have achieved impressive performance in quasi-static manipulation, but struggle in dynamic manipulation tasks because they operate on a single observation at inference time. We identify two representational failures that underlie this limitation. The first is \emph{motion ambiguity}, where a single observation does not include scene dynamics and therefore cannot anticipate the future state of moving objects. The second is \emph{state aliasing}, where visually similar observations from different points in a task require different actions. We argue that these failures persist regardless of model scale and inference latency, showing that the bottleneck is missing temporal context rather than model capacity. Based on this insight, we propose \ours, which augments a pretrained VLA with two temporal inputs: a motion summary extracted from a frozen video foundation model to resolve motion ambiguity and a compact proprioceptive history to resolve state aliasing. \ours\ requires no modification to the backbone and adds minimal compute overhead at training or deployment. Across four dynamic manipulation tasks, it improves Bottle Handover success from 44\% to 74\% and is the only method that solves state aliasing. Probing and ablation studies confirm that each temporal signal independently addresses its corresponding failure. We further release \benchname, a benchmark of over 50k annotated frames for evaluating motion-aware robot perception in both regression and multiple-choice formats.
Project Website: \url{https://tempo-robot.github.io/}

    \vspace{-0.75em}
\end{abstract}

\keywords{VLA Models; Dynamic manipulation; Representation Learning}

\section{Introduction}
\label{sec:introduction}
\begin{figure}[t]
	\centering
	\includegraphics[width=\columnwidth]{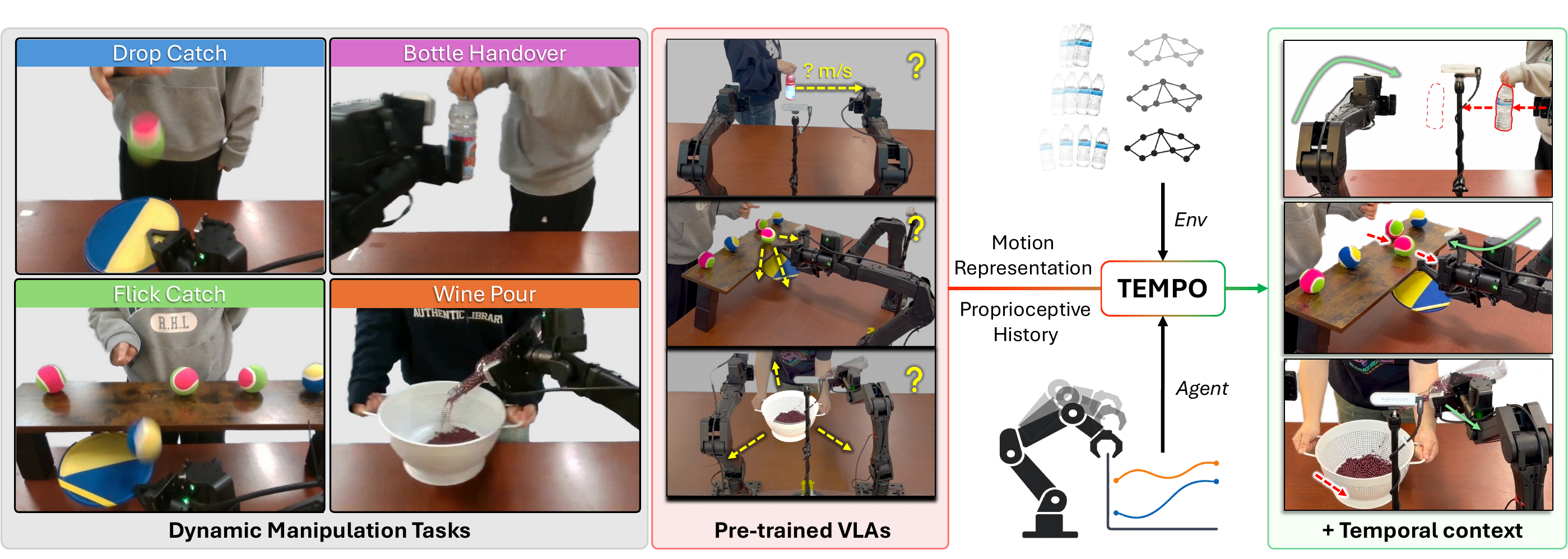}
	\caption{\textbf{\ours\ makes a pretrained VLA motion-aware.}
	Four dynamic tasks (\emph{left}) in which the target object
	is in motion throughout the episode.
	A standard VLA, conditioned on a single observation, cannot determine
	how the scene is evolving (\emph{center, red}; ``?\,m/s'') and cannot
	disambiguate perceptually identical observations that arise at
	different phases of the same task, producing failed grasps and
	mistimed actions.
	\ours\ resolves both failure modes by augmenting the policy with two
	compact signals: \TEMPOMOT, a motion representation from a frozen
	video foundation model, and \TEMPOACT, a compact proprioceptive
	history providing subtask context.
	Together they enable reliable task completion across all four tasks
	(\emph{right, green}).}
	\label{fig:teaser}
\end{figure}

Vision-language-action models have demonstrated strong generalization on quasi-static manipulation, where objects remain at rest throughout the inference interval and a single-frame observation provides a sufficient basis for action~\cite{pi0, pi05, openvla}. This premise does not hold when objects move. A robot intercepting a bottle from a person walking past must anticipate where the bottle will be when the grasp executes, not merely where it was observed at inference time. A single frame samples the scene's configuration but contains no information about how it is evolving. Reducing inference latency narrows but does not close this gap. Even with the freshest possible observation, a snapshot encodes the configuration of the scene, not motion~(\figref{fig:teaser}). We refer to this failure mode as \emph{motion ambiguity}.

A distinct failure mode arises from task structure. Tasks that unfold over sequential subtasks produce perceptually similar frames at qualitatively different points: a hand approaching at the start of a grasp and the same hand receding at its conclusion can appear nearly identical, yet require opposite actions. A policy conditioned only on the current frame cannot disambiguate them, producing jitter and mistimed commitments even on tasks it can nominally complete. We refer to this failure mode as \emph{state aliasing}. Both failures follow from the Markovian assumption shared by every pretrained VLA today. \cref{sec:diagnosis} shows each empirically on current baselines and identifies the temporal context each requires.

We present \ours~(\figref{fig:teaser}), which addresses both failure modes via two compact temporal signals that build on representations already in wide use, without modifying the pretrained backbone. The first, \TEMPOMOT, is a motion representation derived from a frozen video foundation model. Such models, trained on dense visual correspondence, separate scene dynamics from appearance without task-specific supervision, providing the temporal information missing from a single frame. The second, \TEMPOACT, is a compact summary of the robot's own recent proprioceptive history, providing a task-agnostic cue for which subtask is in progress and distinguishing observations that would otherwise appear identical at different task phases. Both signals require no per-task engineering and together add only 2M parameters (0.08\% overhead), leaving the pretrained backbone entirely intact. Asynchronous inference~\cite{rtc, vlash} addresses a complementary bottleneck of execution latency; we build \ours\ on top of an asynchronous backbone~\cite{vlash} so the gain over the async-only baseline isolates the representational contribution.

We evaluate on a bimanual platform across four dynamic tasks designed to expose each failure mode in isolation and combination: Bottle Handover, Drop Catch, Flick Catch, and Wine Pour. On tasks where motion perception or state disambiguation is the limiting factor, \ours\ outperforms both asynchronous baselines: Bottle Handover improves from 44\% to 74\%, and \ours\ is the only system to complete the full versions of Wine Pour and Flick Catch, where both baselines score $0\%$ due to unresolved state aliasing. Results on Drop Catch further validate the framework: it is the one task where execution timing dominates over representation, and it is precisely where asynchronous inference shines, consistent with the complementary relationship we describe above~(\figref{fig:jittery}). Ablations and probing analyses confirm that each signal is individually necessary, relied upon at distinct moments (\TEMPOMOT\ during tracking, \TEMPOACT\ at commitment), and that the motion token encodes object velocity.

We argue that extending pretrained VLAs to dynamic settings is primarily a question of representation, not scale: the temporal context needed to act in a moving scene decomposes into two distinct signals, each tied to a distinct failure mode and independently verifiable. Our contributions are: (1) we identify and formalize two representational failure modes of single-frame VLAs in dynamic settings, \emph{motion ambiguity} and \emph{state aliasing}, and show that neither is a latency nor a capacity failure; (2) we address such failure modes and introduce two minimal plug-and-play inputs, \TEMPOMOT, an abstract past motion representation, and \TEMPOACT, a compact proprioceptive history, that together make any pretrained VLA motion-aware without extensive pretraining efforts; and (3) we release \benchname, an object-motion benchmark of over 50k frames of human-robot dynamic manipulation with per-frame object-velocity annotations, together with a multiple-choice variant in the format of MVBench~\cite{li2024mvbench} and VLM4D~\cite{zhou2025vlm4d} with object moving direction and motion magnitude question, targeted at evaluating motion understanding in VLMs and VLAs.

\section{Related Work}
\label{sec:related-work}

\noindent\textbf{Dynamic Tasks for Robotics.}
Dynamic manipulation has a long research history, with compelling results on table tennis~\cite{tabletennis, mstennis}, object tossing~\cite{zeng2020tossingbot, flingbot}, and aerial catching~\cite{catchit, dynamichandover}. These systems achieve strong performance by explicitly modeling object physics and designing task-specific controllers, but each is an end-to-end pipeline for a single scenario that does not transfer to other tasks. In contrast, pretrained VLA models~\cite{brohan2022rt, driess2023palm, openvla, rt2, octo, pi0, pi05} generalize across tasks without per-task engineering; extending them to dynamic settings is the focus of this work.

\noindent\textbf{Asynchronous Inference for Dynamic Tasks.}
A prominent line of work addresses the \emph{execution timing} problem: inference latency of VLA models causes the executed chunk~\cite{chi2023diffusion,zhao2023act} to be conditioned on a stale observation. SmolVLA~\cite{smolvla}, RTC~\cite{rtc}, VLASH~\cite{vlash}, and Leave No Observation Behind~\cite{leave_no} each reduce or correct for this staleness. DynamicVLA~\cite{dynamicvla} co-designs a compact VLA architecture with continuous inference and latent-aware streaming for lower-latency dynamic control. F2F-AP~\cite{f2fap} takes a different route within the same problem: it predicts optical flow to synthesize an anticipated future frame, giving the policy visual context that compensates for latency. \ours\ is orthogonal to all of these: asynchronous inference addresses \emph{when} a prediction is applied; \ours\ addresses \emph{what} the policy is conditioned on. Even with a fresh or anticipated observation, a snapshot cannot encode how the scene is evolving or disambiguate sequential task phases. We build \ours\ on the asynchronous backbone of~\cite{vlash} so our experiments directly isolate the contribution of motion-aware representation; the two approaches are complementary and can be combined.

\noindent\textbf{Visual and Visuomotor Representations in Robot Learning.}
A parallel line of work asks what \emph{static} representation a policy should be conditioned on. Visual pretraining from large-scale ego-centric or robot video~\cite{r3m, mvp, sensorimotor} learns transferable features~\cite{data4robotics} before any downstream policy fine-tuning. RoboAffordances~\cite{roboaffordances} instead distills task-agnostic affordance priors from human videos, and HPT~\cite{hpt} scales pretraining across heterogeneous embodiments. UVT~\cite{uvt} and Anchor-Align~\cite{anchoralign} shape the training \emph{signal} rather than the input. All of these methods condition the policy on a single frame. \ours\ instead supplying the temporal context that no single frame can carry.


\section{Motion Ambiguity and State Aliasing}
\label{sec:diagnosis}

As briefed in \cref{sec:introduction}, we observe two dominant failure modes recent VLA encounters when trained on dynamic tasks, both stemming from a Markovian assumption shared by every pretrained VLA today, where the current observation is treated as a sufficient context for inferring action. We illustrate each on our four dynamic tasks (setup in \cref{sec:exp-setup}).

\begin{figure}[!t]
	\centering
	\includegraphics[width=0.9\columnwidth]{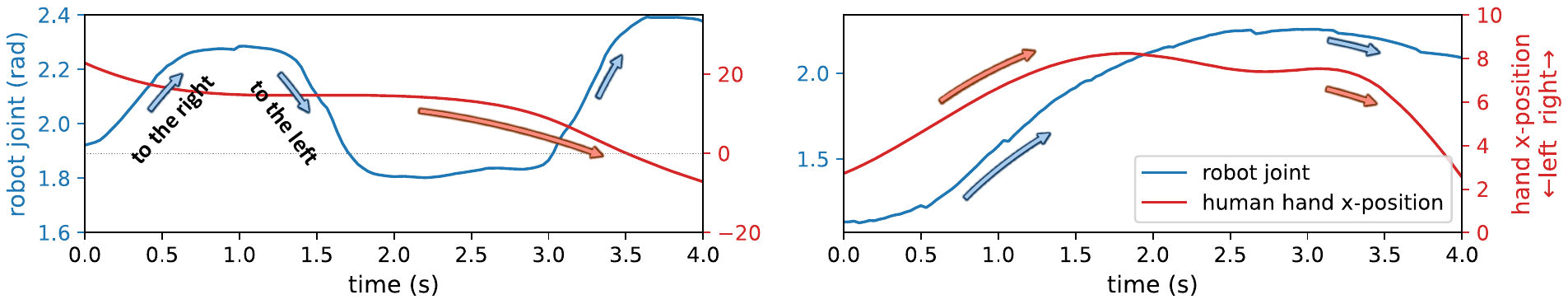}
	\vspace{-1em}
	\caption{\textbf{Robot joint action (blue) does not track the human hand (red).} Side-by-side comparison of a Drop Catch rollout from the baseline (VLASH, left) and \ours~(right). The baseline oscillates left-to-right continuously while the human hand moves in a single direction, whereas \ours~closely tracks the hand.}
	\label{fig:jittery}
\end{figure}

\noindent\textbf{Motion ambiguity.}
A policy $\pi(o_t)$ conditioned on the current frame observes where objects are, but not where they are going. Recent works on dynamic tasks all focus on reducing inference latency, which lets $\pi$ act on a more recent observation, but does not inform the policy of object motion. \figref{fig:jittery} shows the resulting behavior on Drop Catch: when the policy is supposed to shadow human hand motion, the VLASH~\cite{vlash} policy instead produces actions that oscillate left-to-right. The same representational gap is also evident in our motion probe study in \cref{para:motion_probe}: linearly decoding object velocity from each baseline collapses on tasks where object motion is the only cue, with $R^2 \le 0.03$ on Flick Catch and $\le 0.09$ on Wine Pour (\cref{tab:motion_probe}). The policy has no basis for anticipating where a moving target will be.

\begin{wraptable}{r}{0.45\columnwidth}
\centering
\small
\setlength{\tabcolsep}{4pt}
\begin{tabular}{lcc}
\toprule
Method & Flick Catch & Wine Pour \\
\midrule
RTC~\cite{rtc} & 0\% & 0\% \\
VLASH~\cite{vlash} & 0\% & 0\% \\
\ours~(Ours) & \textbf{68\%} & \textbf{97.6\%} \\
\bottomrule
\end{tabular}
\caption{\textbf{Success on the untrimmed Flick Catch and Wine Pour}, where each episode retains the release-and-retract phase that produces state aliasing. Both asynchronous baselines fail every rollout; \ours\ completes the task by using \TEMPOACT\ to disambiguate phases the current frame cannot.}
\label{tab:state_alias}
\end{wraptable}

\noindent\textbf{State aliasing.}
A second failure mode arises from task structure and persists in the absence of motion. Multi-stage tasks force the policy to produce opposite actions from visually near-identical moments. For example, reaching for a plate at the start of a task and releasing it and pulling the gripper back look nearly the same from the head camera, but the corresponding future actions are opposite. The policy then hesitates between the two options: the gripper commits to neither, and the task fails to advance. Trained on the original Wine Pour and Flick Catch tasks, both asynchronous baselines fail every rollout ($0/50$; \cref{tab:state_alias}). To obtain nonzero baseline numbers for the main results in \cref{tab:success_rates}, we manually remove aliased state by trimming the release-and-retract phase from every training episode. Having to remove this segment at all is itself evidence of the failure for our baseline models.

\noindent\textbf{What the two failures require.}
Both failures come down to the same gap: a single frame carries no temporal context. To perceive scene motion, the policy needs a signal built from recent observations; to tell task phases apart, it needs a signal built from its own recent actions. Both signals summarize data the policy is already receiving at inference time.


\section{\methodname}
\label{sec:method}

\begin{figure}[t]
	\centering
	\includegraphics[width=0.95\columnwidth]{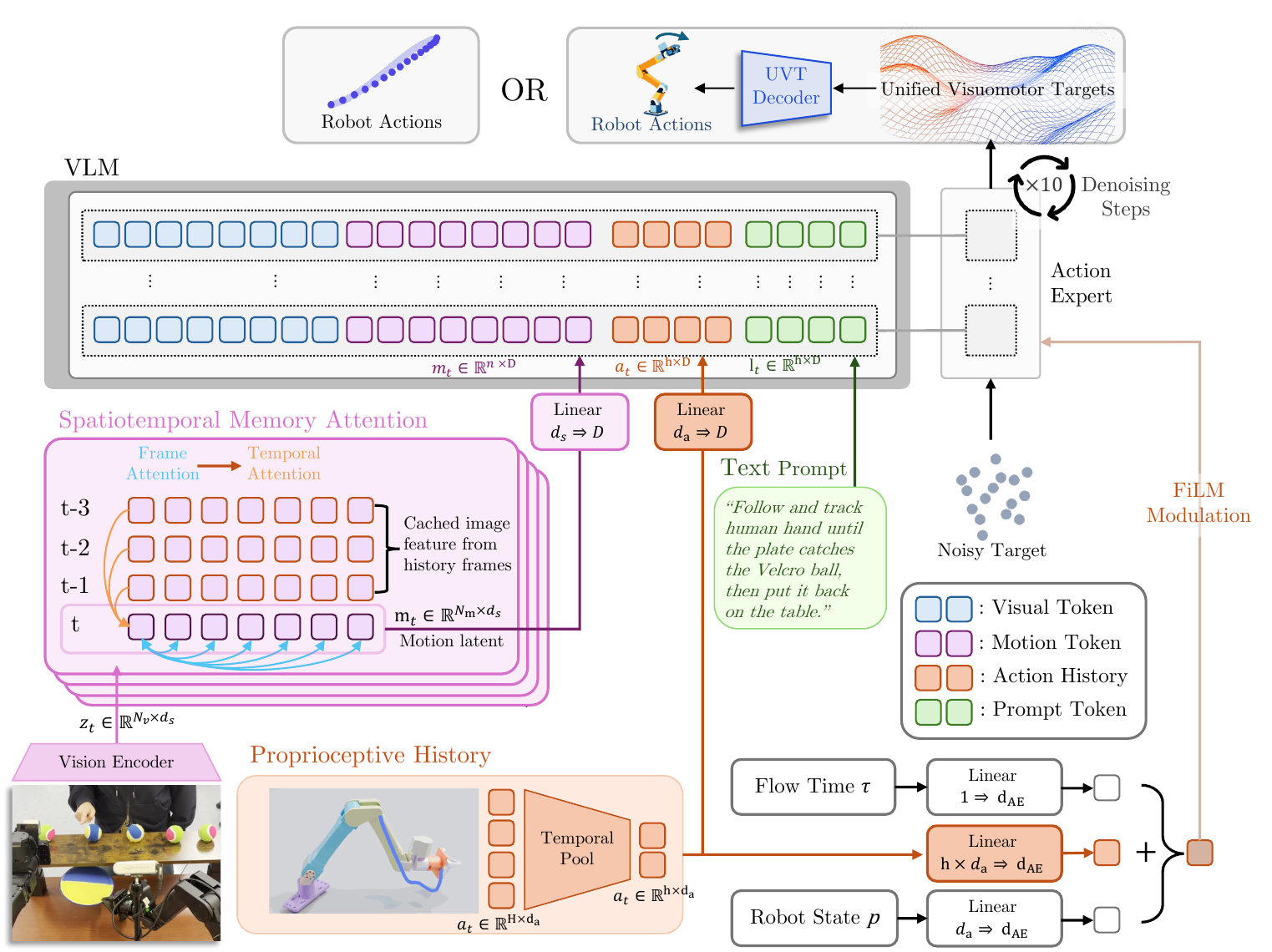}
	\vspace{-0.6em}
	\caption{\textbf{\ours\ architecture.} \ours\ augments a pretrained VLM-based policy with two temporal inputs. \TEMPOMOT\ cross-attends the current frame's patch features $\mathbf{f}_t$ against a streaming cache $\mathcal{F}_t$ of features from the preceding $W-1$ frames, producing a motion code $\mathbf{m}_t$ that encodes scene dynamics. \TEMPOACT\ mean pools the robot's recent proprioceptive command history into $K$ fixed-size buckets, yielding a compact summary $\mathbf{c}_t$. Both signals attach to the pretrained backbone with near-zero overhead: $\mathbf{m}_t$ and $\mathbf{c}_t$ are projected to the VLM's token dimension and appended to its prefix, and $\mathbf{c}_t$ additionally conditions the action expert via an AdaRMS residual.}
	\label{fig:method}
\end{figure}

\ours\ supplies the two temporal signals that \cref{sec:diagnosis} identifies as missing: \TEMPOMOT, a motion encoder that summarizes recent scene dynamics, and \TEMPOACT, a compact proprioceptive history. Both attach to a pretrained VLA with near-zero overhead (\figref{fig:method}).

\vspace{-0.3em}
\subsection{\TEMPOMOT: Captures Object and Scene Dynamics}
\TEMPOMOT\ summarizes recent visual dynamics by cross-attending the current frame against a rolling cache of past frames (\figref{fig:method}). Let $\mathbf{f}_\tau$ denote the patch features of observation $o_\tau$, extracted by $\phi$'s per-frame encoder. Over a streaming window of length $W$, we maintain a cache $\mathcal{F}_t = \{\mathbf{f}_\tau\}_{\tau=t-W+1}^{t-1}$ of features from the preceding $W-1$ frames, populated incrementally so that each $\mathbf{f}_\tau$ is computed exactly once and reused on subsequent steps at no additional cost. The motion code is then
\begin{equation}
\label{eq:tempomot}
\mathbf{m}_t \;=\; \phi(o_{t-W+1:t}) \;=\; \operatorname{CrossAttn}\bigl(\mathbf{f}_t,\; \mathcal{F}_t\bigr),
\end{equation}
with the current-frame features $\mathbf{f}_t$ as queries and historical visual features as keys and values.

\begin{wraptable}{r}{0.42\columnwidth}
\centering
\vspace{-\baselineskip}
\small
\setlength{\tabcolsep}{4pt}
\begin{tabular}{lc}
\toprule
Motion encoder $\phi$ & Handover \\
\midrule
SAM~2.1-Tiny~\cite{sam2} (default) & \textbf{74\%} \\
Reducio~\cite{reducio}       & \textbf{76\%} \\
VidTwin~\cite{vidtwin}       & 60\% \\
VideoLaVIT~\cite{videolavit} & 52\% \\
\bottomrule
\end{tabular}
\caption{\textbf{Swapping the motion encoder $\phi$} keeps Bottle Handover success comfortably above the VLASH baseline ($38\%$).}
\label{tab:encoder_swap}
\end{wraptable}
We instantiate $\phi$ with a frozen pretrained video foundation model, used off-the-shelf without motion-specific fine-tuning; SAM~2.1-tiny~\cite{sam2} is our default. The choice is flexible: any video encoder whose features separate motion from appearance can serve as $\phi$. \figref{fig:ball_velocity_probe} shows that object velocity is linearly decodable from $\mathbf{m}_t$ computed from different video encoders, substantially outperforming a position-only baseline, across several encoder families, and \cref{tab:encoder_swap} confirms the same conclusion at the task level: swapping $\phi$ across four video foundation models keeps Bottle Handover success in the $52$–$76\%$ range, all well above the VLASH baseline. Reducio slightly exceeds our SAM default; we still adopt SAM~2.1-Tiny for the streaming speed required at real-time control rates (\tabref{tab:latency_comparison}).

\vspace{-0.3em}
\subsection{\TEMPOACT: Resolves State Aliasing}
Where \TEMPOMOT\ captures how the scene is moving, \TEMPOACT\ captures what the robot has been doing. A compact summary of recent proprioceptive history gives the temporal context for the robot policy to commit to the correct task phase. Without resolving state aliasing, the policy averages over multiple valid actions for the same observation, often producing indecisive behavior (e.g., pausing) or oscillating between conflicting actions, as seen in our baseline rollouts (\cref{fig:ambiguity}).

Raw proprioceptive history is long, high-frequency, and largely redundant, so we compact it into a fixed-size summary $\mathbf{c}_t$ (\TEMPOACT). Let the proprioceptive command at time $\tau$ be $q_\tau\in\mathbb{R}^{d_q}$. We consider the most recent $H$ commands preceding time $t$, partition them into $K$ contiguous temporal buckets $B_1,\ldots,B_K$ of equal length $L$ (so that $H=KL$), ordered from oldest to newest, and summarize each bucket by its mean:
\begin{equation}
\label{eq:joint-compaction}
\mathbf{c}_t=\bigl(\bar q^{(1)}_t,\dots,\bar q^{(K)}_t\bigr),\qquad
\bar q^{(k)}_t=\frac{1}{L}\sum_{\tau\in B_k} q_\tau\in\mathbb{R}^{d_q},\qquad
q_\tau:=0\ \text{for }\tau<\text{episode start}.
\end{equation}

We use $K=10$ buckets, each averaging $L=30$ consecutive commands. Buckets that fall entirely before the episode start are marked with a pad flag. The resulting representation $\mathbf{c}_t\in\mathbb{R}^{K\times d_q}$ always contains $K$ tokens, regardless of how much history it summarizes. Extending the temporal horizon only changes the bucket length $L$, not the number of tokens. This simple, parameter-free aggregation preserves a coarse trend of robot's recent motion while adding negligible computational overhead.

\vspace{-0.3em}
\subsection{Plug-and-Play Temporal Integration}
We add the two signals to a pretrained VLA while keeping the pretrained backbone largely intact (\figref{fig:method}). We build on VLASH's asynchronous-inference pipeline~\cite{vlash} as the backbone, though our additions are agnostic to this choice. $\mathbf{m}_t$ is projected to the VLM's token dimension and appended to its prefix alongside the current multi-view observation and the language instruction. The action expert reads the resulting VLM prefix through the cross-attention layers. 

The $K$ bucket tokens of $\mathbf{c}_t$ are used in two ways. First, they are projected into VLM's prefix tokens dimension and appended to the VLM token stream, allowing the multimodal prefix to include the robot's recent proprioceptive history. Second, the same bucket sequence is flattened and passed through a conditioning MLP to produce an AdaRMS residual for the action expert, modulating the denoising process directly. 

Together, the two temporal signals provide complementary temporal contexts to the VLA.
The added conditioning path is zero-initialized, so the pretrained policy's behavior is preserved at the start of fine-tuning. Flow-matching loss is unchanged, and \ours\ adds only 2M~($0.08$\%) parameters.

\vspace{-0.3em}
\subsection{Real-Time Deployment via Parallel Encoding}
\begin{wraptable}{r}{0.40\columnwidth}
\centering
\vspace{-\baselineskip}
\small
\setlength{\tabcolsep}{4pt}
\begin{tabular}{lc}
\toprule
\textbf{Method} & \textbf{Median} \\
\midrule
VLASH~\cite{vlash}          & 33.8 ms \\
RTC~\cite{rtc}              & 127.7 ms \\
\ours~(Ours)                & 35.3 ms \\
\bottomrule
\end{tabular}
\caption{Inference latency comparison across methods.}
\label{tab:latency_comparison}
\end{wraptable}
\TEMPOMOT\ is encoded in a dedicated background thread, decoupled from the control loop: the motion encoder consumes camera frames at sensor rate ($\sim$11\,ms per frame on an NVIDIA RTX PRO 6000) and continuously updates the latest motion token, which the policy reads at each control step. The only overhead the policy pays is about 1.5\,ms for attending to the extra token (35.3\,ms vs.\ 33.8\,ms median forward time; \tabref{tab:latency_comparison}), versus approximately 128\,ms per chunk for the asynchronous baseline RTC~\cite{rtc}. Because $\phi$ operates over a fixed-size streaming window, its per-frame cost stays constant regardless of history length, so \ours\ adds motion awareness without impacting the policy's control rate.


\section{Experiments}

\begin{table}[h]
\centering
\small
\begin{tabular}{lccccc}
\toprule
Method & Bottle Handover & Drop Catch & Flick Catch & Wine Pour & Average \\
\midrule
RTC~\cite{rtc} & 44\% & 80\% & 48\% & 80.8\% & 63.2\% \\
VLASH~\cite{vlash} & 38\% & 38\% & 20\% & 94.0\% & 47.5\% \\
\ours~(Ours)  & 74\% & 66\% & 66\% & \textbf{98.1\%} & 76.0\% \\
TEMPO+UVT~\cite{uvt} & \textbf{88\%} & \textbf{88\%} & \textbf{76\%} & 96.7\% & \textbf{87.2\%}\\
\bottomrule
\end{tabular}
\vspace{1em}
\caption{\textbf{Success rates across the four dynamic tasks.} Each
entry is the percentage of $50$ real-world rollouts in which the policy completes
the task, evaluated under varied object speeds, appearances,
and approach directions. For Wine Pour task, instead of recording binary success or failure, we measure the percentage of the "wine" mass retained in the pot at the end of the pour.}
\label{tab:success_rates}
\vspace{-4mm}
\end{table}

\label{sec:experiments}

\subsection{Experimental Setup}
\label{sec:exp-setup}

\label{sec:setup}
\begin{figure}[!tbp]
	\centering
	\includegraphics[width=0.5\columnwidth]{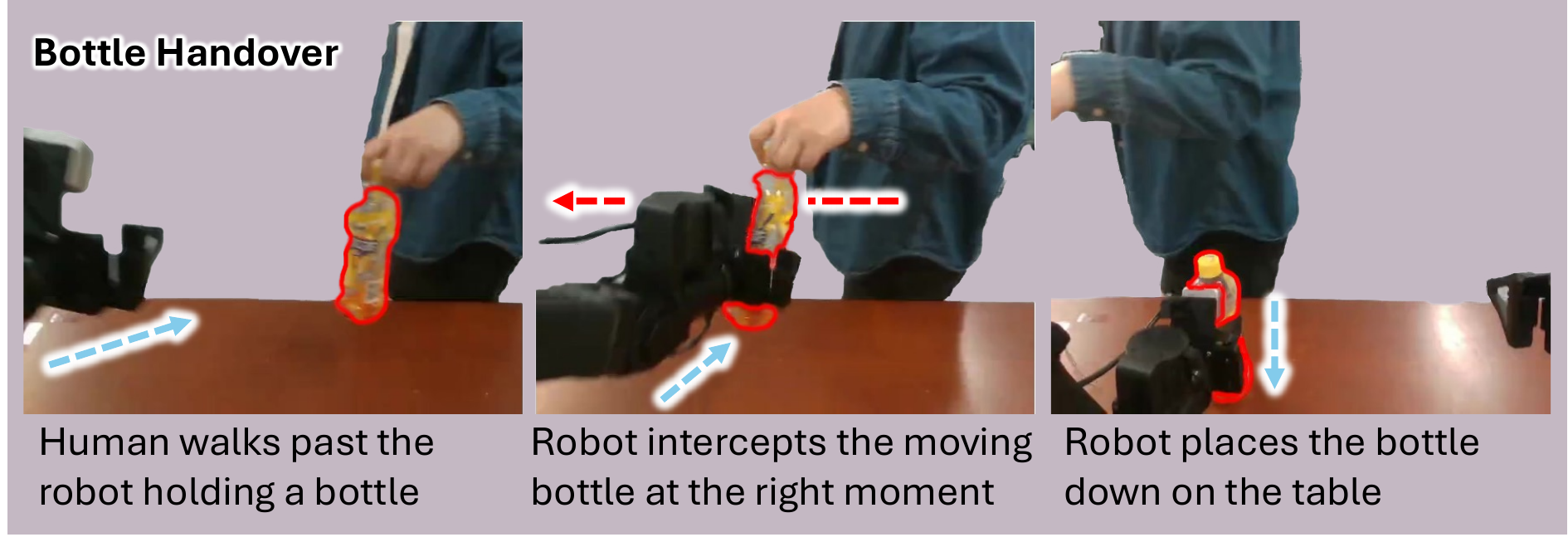}\hfill
	\includegraphics[width=0.5\columnwidth]{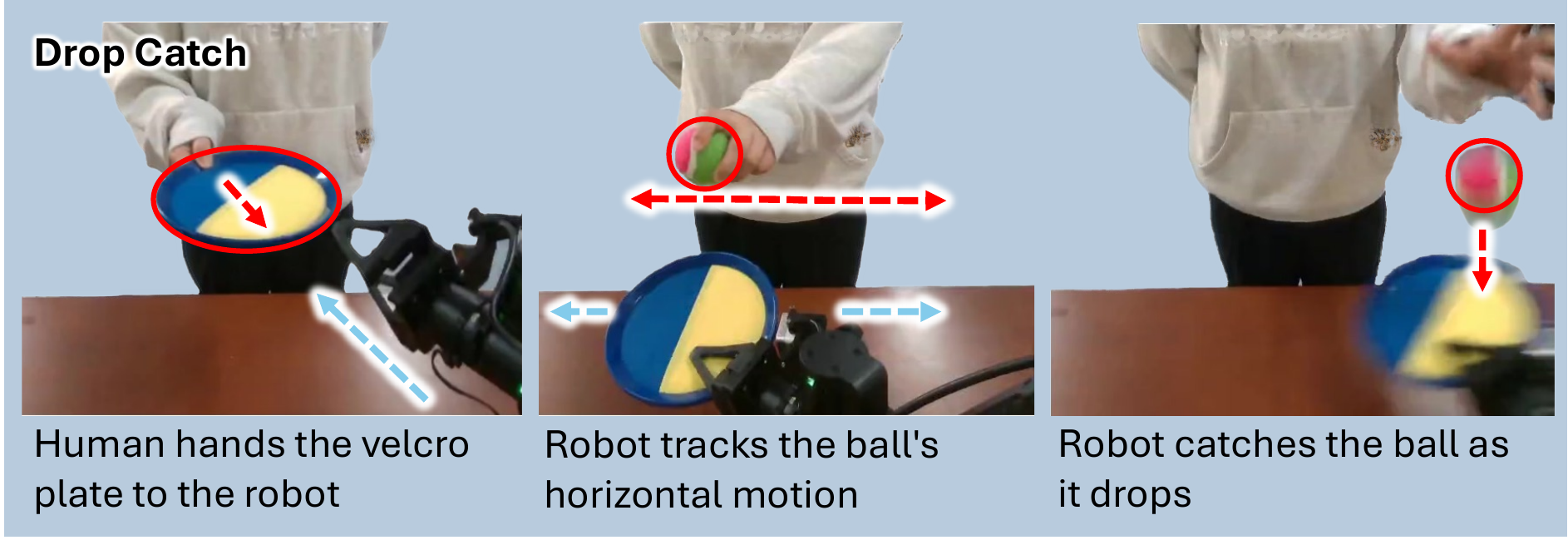}\\
	\includegraphics[width=0.5\columnwidth]{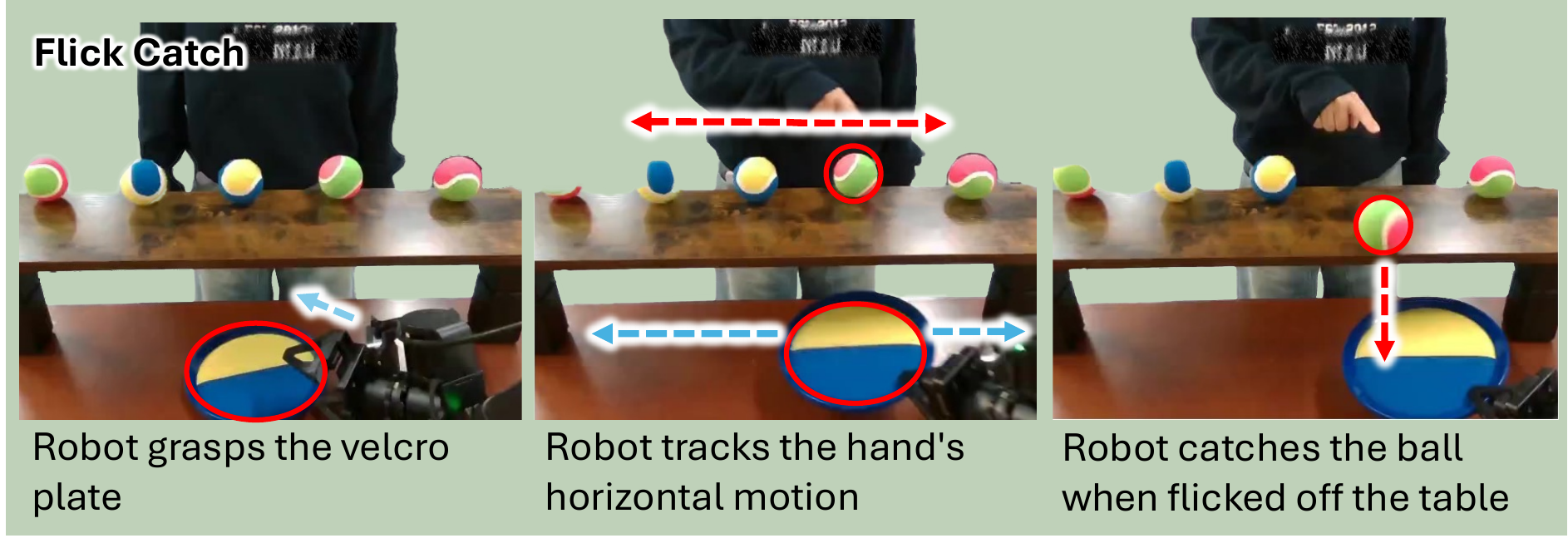}\hfill
	\includegraphics[width=0.5\columnwidth]{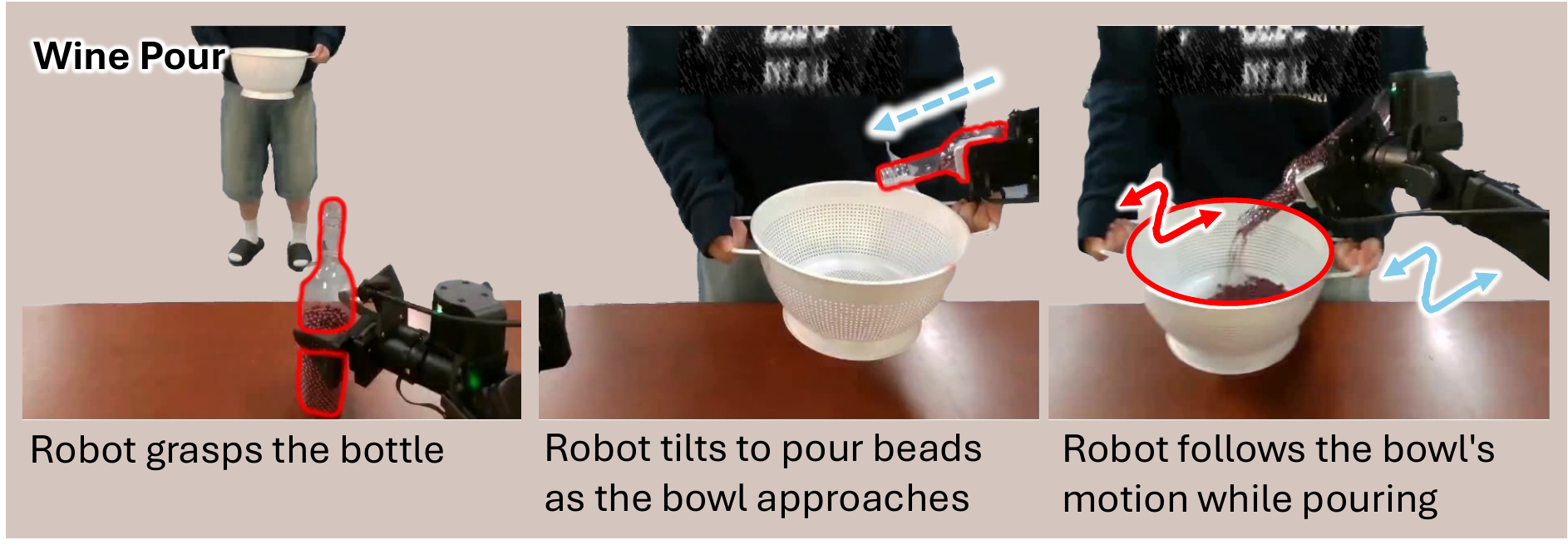}
	\caption{\textbf{Task Setup.} We benchmark all models across four comprehensive dynamic tasks.}
	\label{fig:tasks}
\end{figure}

We design four tasks to expose the two ambiguities identified in \cref{sec:introduction}: motion ambiguity and state aliasing. We ask whether \ours\ resolves each kind of ambiguity through \TEMPOMOT\ and \TEMPOACT. We compare \ours\ against two state-of-the-art asynchronous baselines designed for responsive task execution, RTC~\cite{rtc} and VLASH~\cite{vlash}. For Wine Pour and Flick Catch, both baselines collapse to $0\%$ on the full task because the release-and-retract phase is perceptually aliased with reaching. As a result, a baseline that sees the gripper near the bottle or plate cannot tell whether to grasp or withdraw and oscillates between the two (more detailed visualization and analysis in \cref{app:failure}). To avoid a trivial comparison, we report quantitative results on a trimmed version that removes the release-and-retract phase, eliminating this artifact. We emphasize that \ours\ disambiguates these phases through \TEMPOMOT\ and \TEMPOACT, so the gain in \cref{tab:success_rates} is a lower bound on \ours's full-task advantage. For qualitative rollouts of the baselines' complete failure and \ours's behavior, please see our project webpage and \cref{app:failure}.

\noindent\textbf{Task properties.}\label{sec:task_properties} Each of the four tasks is dynamic and shares four properties that a current-frame observation alone cannot resolve.
  \textbf{(P1)} The target object is moving at varying motion. Therefore, reacting based on the current snapshot alone is inaccurate in both position and timing.
  \textbf{(P2)} The relevant target shifts during the episode. As a result, the policy must redirect its attention to whichever object currently matters.
  \textbf{(P3)} Each task decomposes into a sequence of subtasks (approach, grasp, track, release), and the correct next action depends on which subtask has already been completed.
  \textbf{(P4)} The policy faces state ambiguity. From the current frame alone, multiple next actions are plausible, and a perceptually identical state may demand opposite actions depending on what has come before.

\textbf{Bottle Handover (P1).}~The robot intercepts a bottle from a human walking by from either direction. We vary the walking speed, bottle's height, and holding pose across episodes, so the policy must estimate the bottle's velocity to time and place the grasp correctly.
\\\textbf{Drop Catch (P1, P3).}~The robot catches a ball dropped by a horizontally moving human hand. Each episode begins with the human handing a Velcro plate to the robot, after which the robot tracks the human’s hand laterally before catching the released ball. This task includes a moving dynamic object (P1) and multiple sequential subtasks (P3).
\\\textbf{Flick Catch (P1--P4).}~Several Velcro balls rest on a long horizontal platform. The robot first picks up a Velcro plate from the table (P3). A human hand then hovers over the balls and, at a random time, flicks one off the platform. The robot must first track the hand, then switch to tracking the released ball to catch it in time (P2, P1). Because the robot picks up and later places down the plate at the same location, the pick-up is perceptually aliased with the subsequent drop-off (P4), making this the most challenging of the four tasks.
\\\textbf{Wine Pour (P1--P4).}~The robot pours beads, an alternative to wine, into a pot that the human moves around. The task decomposes into three subtasks (P3): pick up the bottle from the table, pour while tracking the pot movement, and put the bottle back down. Pick-up and put-down produce visually similar states yet require opposite actions, producing state aliasing (P4). Once pouring begins, the target shifts from the bottle to the pot (P2), whose motion must be tracked (P1). Since every rollout inevitably spilled some beads, this task's success is determined by the percentage of beads' mass retained in the pot at the end of the pour.

\subsection{Experimental Results}
\label{sec:quant}

\begin{figure}[!tbp]
	\centering
 	\includegraphics[width=0.8\columnwidth]{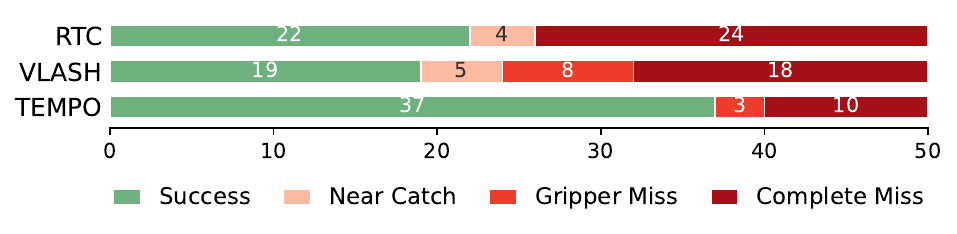}
    \vspace{-1em}
	\caption{\textbf{Failure-mode breakdown on Bottle Handover}, we categorized and visually examined each episode in deployment rollouts and attributed each model's failure to different categories.}
	\label{fig:failure_modes}
\end{figure}

\noindent\textbf{Success Rate on Dynamic Tasks.}
As shown in \cref{tab:success_rates}, we extensively compare our method against each of the SOTA asynchronous VLA models. For each task, we calculated success rate for each model based on 50 trials following the evaluation protocol established under \cref{sec:setup}. On simpler tasks like Drop Catch, \ours\ achieves a competitive success rate compared to RTC, while VLASH frequently overshoots the human motion and misses the dropped ball. Across the remaining tasks where the object motion is more uncertain or the object of interest shifts, \ours\ consistently outperforms both baselines, including $+36\%$ on Bottle Handover, $+4.1\%$ on Wine Pour, and $+46\%$ on Flick Catch. We observed closer tracking of the human hand by the robot gripper in Drop Catch and Flick Catch, as well as more precisely timed grasps in Bottle Handover.

\noindent\textbf{Failure mode analysis.}
A single success rate hides \emph{why} a policy fails, so we visually examined all
$50$ rollouts of every model on Bottle Handover and labeled each as a success or
one of three failure modes (\figref{fig:failure_modes}). The dominant failure for both baselines is \emph{Complete Miss}, where the arm reaches toward the wrong place and misses the bottle entirely, indicating that the policy mispredicts where the moving bottle will be. VLASH and RTC commit $18$ and $24$ Complete Misses respectively, while the timing-related \emph{Gripper Miss} and the near-success \emph{Near Catch} are comparatively rare. By conditioning on recent motion, \ours\ cuts Complete Misses to $10$ and raises success from $19$ (VLASH) and $22$ (RTC) to $37$ out of $50$. Although both asynchronous pipelines significantly reduce inference delay, a lack of motion understanding still hurts dynamic-task performance.

\begin{wraptable}{r}{0.42\columnwidth}
\centering
\vspace{-\baselineskip}
\small
\setlength{\tabcolsep}{4pt}
\begin{tabular}{lcc}
\toprule
VLA backbone & Base & + \ours \\
\midrule
$\pi_{0.5}$~\cite{pi05}   & 26\% & \textbf{56\%} \\
RTC~\cite{rtc}            & 44\% & \textbf{80\%} \\
VLASH~\cite{vlash}        & 38\% & \textbf{74\%} \\
\bottomrule
\end{tabular}
\caption{\textbf{\ours\ is plug-and-play across VLA backbones.} Adding \TEMPOMOT\ and \TEMPOACT\ improves Bottle Handover on all backbones, including the non-asynchronous $\pi_{0.5}$.}
\label{tab:backbone_integration}
\end{wraptable}
\noindent\textbf{Plug-and-play across VLA backbones.}
The two temporal inputs of \ours\ are not backbone-specific. We attach the same \TEMPOMOT\ and \TEMPOACT\ to two additional pretrained VLAs, $\pi_{0.5}$~\cite{pi05} (non-asynchronous) and RTC~\cite{rtc}, and retrain each on the same Bottle Handover demos. Every backbone we tried improves substantially: $\pi_{0.5}$ from $26\%$ to $56\%$, RTC from $44\%$ to $80\%$, and VLASH from $38\%$ to $74\%$ (\cref{tab:backbone_integration}). \ours\ helps both asynchronous and non-asynchronous backbones, indicating that motion-aware perception is a general upgrade rather than a VLASH-specific effect.

\noindent\textbf{Extending TEMPO with a compact training target.}
With \ours\ closing the input-side gap, a natural line of question being whether anything remains on the output side. Our prior work, Unified Visuomotor Target (UVT)~\cite{uvt}, argues that it does: VLAs are typically trained to predict raw joint-position chunks, which are high-dimensional and jitter-prone, and swapping this target for a compact motion primitive aligned with object motion tightens the target the policy has to learn. Since \ours\ and UVT act on different sides of the policy, they combine without redundancy: \ours+UVT achieves the highest overall success rate in \cref{tab:success_rates} and outperforms baselines on $3$ of the $4$ tasks, further pushing dynamic task performances.

\section{Analysis and Ablations}
\label{sec:ablation}

Having shown \ours\ outperforms recent VLA baselines on dynamic tasks, we now turn to \emph{why}. We evaluate whether the policy relies on \TEMPOMOT\ and \TEMPOACT\ at the task stages where their corresponding ambiguities arise, using single-signal ablations and hidden-state probing.

\begin{wraptable}{r}{0.38\columnwidth}
\centering
\vspace{-\baselineskip}
\small
\setlength{\tabcolsep}{4pt}
\begin{tabular}{lc}
\toprule
Configuration & Handover \\
\midrule
\TEMPOACT\ only & 52\% \\
\TEMPOMOT\ only & 56\% \\
\ours\ (full)   & \textbf{74\%} \\
\bottomrule
\end{tabular}
\caption{\textbf{Both temporal signals contribute to the task.} Removing either \TEMPOMOT\ or \TEMPOACT\ reduces Bottle Handover success; the two together outperform either alone.}
\label{tab:input_ablation}
\end{wraptable}
\noindent\textbf{Each temporal signal is individually necessary.}
To verify that both \TEMPOMOT\ and \TEMPOACT\ carry non-redundant information for task success, we retrain \ours\ with each single signal removed and evaluate on Bottle Handover (\cref{tab:input_ablation}). Removing either signal substantially reduces success, and both single-signal variants still outperform the VLASH baseline (38\%), confirming that each temporal input is independently useful. Combining the two lifts success to $74\%$, above what either signal alone achieves. Each signal is also consulted at a distinct moment in the episode; see \cref{app:input_ablation} for the per-frame ablation MSE.

\noindent\textbf{Baseline hidden states lack motion on the hardest tasks.}
\label{para:motion_probe}
We test whether the trained policy encodes object motion by probing its hidden state at a fixed token position, following the same protocol as prior VLM probe works~\cite{zhao2024first, lu2025probing}. For each task we annotate the per-frame position of the object of interest and compute its pixel-space velocity, yielding an object-motion benchmark of over 50k annotated frames of human-robot dynamic manipulation that we release as \benchname, together with a multiple-choice variant in the format of MVBench~\cite{li2024mvbench} and VLM4D~\cite{zhou2025vlm4d}. Here we report the regression setting: we fit an MLP probe from each model's hidden state to the velocity and report $R^2$ (\cref{tab:motion_probe}).

\ours\ produces the strongest probe result on every row of \cref{tab:motion_probe}. Its hidden state decodes object velocity substantially better than the asynchronous baselines on the dynamic tasks where motion is the only cue (Flick Catch $R^2 = 0.57$ vs.\ ${\le}\,0.03$; Wine Pour $R^2 = 0.44$ vs.\ ${\le}\,0.09$), where the baseline probes collapse to near zero. Since RTC, VLASH, and \ours\ share the same $\pi_{0.5}$ backbone and pretraining, \ours's gain over the two is attributed to its motion-conditioned fine-tuning: \TEMPOMOT\ and \TEMPOACT\ leave a linearly-decodable motion signal in the hidden state that the other $\pi_{0.5}$-family backbones do not carry. We additionally ablate the choice of motion encoder, probing several video foundation models' latents for object velocity, in \cref{app:motion_encoder}.

\begin{table}[!t]
\centering
\small
\begin{tabular}{lccc}
\toprule
Task & RTC~\cite{rtc} & VLASH~\cite{vlash} & \ours \\
\midrule
Bottle Handover  & 0.46 & 0.43 & \textbf{0.52} \\
Drop Catch       & 0.31 & 0.28 & \textbf{0.33} \\
Flick Catch      & $-0.07$ & 0.03 & \textbf{0.57} \\
Wine Pour        & 0.06 & 0.09 & \textbf{0.44} \\
\midrule
Average          & 0.19 & 0.21 & \textbf{0.47} \\
\bottomrule
\end{tabular}
\vspace{1em}
\caption{\textbf{Motion probing of the trained VLA hidden state.} MLP probe on object velocity across our dynamic tasks ($R^2$). On Flick Catch and Wine Pour, where motion is the only usable cue, the asynchronous baselines' representations collapse to near zero while \ours's remains informative.}
\label{tab:motion_probe}
\vspace{-1em}
\end{table}

\label{sec:analysis}





\section{Conclusion}
\label{sec:conclusion}

The gap between pretrained VLAs and dynamic manipulation is primarily representational: motion ambiguity and state aliasing each have a fix from representations already in wide use, and our failure mode breakdown (\figref{fig:failure_modes}) and ablation analyses (\secref{sec:ablation}) confirm that the two failures are separable and that each signal addresses its own. \ours\ implements this as two plug-and-play inputs with near-zero overhead; the gains on the untrimmed Wine Pour and Flick Catch, where asynchronous baselines score zero (\cref{tab:state_alias}), show that timing is not the bottleneck.



\clearpage
\acknowledgments{We thank CoRL reviewers and AC for their valuable feedback on improving papers, along with Basavasagar Patil for his careful reviews of the final draft of our paper.}


\bibliography{example}  

@article{rtc,
  title={Real-time execution of action chunking flow policies},
  author={Black, Kevin and Galliker, Manuel and Levine, Sergey},
  journal={Advances in Neural Information Processing Systems},
  volume={38},
  pages={33383--33407},
  year={2026}
}

@article{vlash,
  title={Vlash: Real-time vlas via future-state-aware asynchronous inference},
  author={Tang, Jiaming and Sun, Yufei and Zhao, Yilong and Yang, Shang and Lin, Yujun and Zhang, Zhuoyang and Hou, James and Lu, Yao and Liu, Zhijian and Han, Song},
  journal={arXiv preprint arXiv:2512.01031},
  year={2025}
}

@article{pi05,
  title={$\pi_{0.5}$: a Vision-Language-Action Model with Open-World Generalization},
  author={{Physical Intelligence} and Black, Kevin and Brown, Noah and Darpinian, James and Dhabalia, Karan and Driess, Danny and Esmail, Adnan and Equi, Michael and Finn, Chelsea and Fusai, Niccolo and Galliker, Manuel Y. and Ghosh, Dibya and Groom, Lachy and Hausman, Karol and Ichter, Brian and Jakubczak, Szymon and Jones, Tim and Ke, Liyiming and LeBlanc, Devin and Levine, Sergey and Li-Bell, Adrian and Mothukuri, Mohith and Nair, Suraj and Pertsch, Karl and Ren, Allen Z. and Shi, Lucy Xiaoyang and Smith, Laura and Springenberg, Jost Tobias and Stachowicz, Kyle and Tanner, James and Vuong, Quan and Walke, Homer and Walling, Anna and Wang, Haohuan and Yu, Lili and Zhilinsky, Ury},
  journal={arXiv preprint arXiv:2504.16054},
  year={2025}
}

@inproceedings{openvla,
  title={OpenVLA: An Open-Source Vision-Language-Action Model},
  author={Kim, Moo Jin and Pertsch, Karl and Karamcheti, Siddharth and Xiao, Ted and Balakrishna, Ashwin and Nair, Suraj and Rafailov, Rafael and Foster, Ethan and Lam, Grace and Sanketi, Pannag and Vuong, Quan and Kollar, Thomas and Burchfiel, Benjamin and Tedrake, Russ and Sadigh, Dorsa and Levine, Sergey and Liang, Percy and Finn, Chelsea},
  booktitle={Conference on Robot Learning},
  year={2024}
}

@inproceedings{rt2,
  title={Rt-2: Vision-language-action models transfer web knowledge to robotic control},
  author={Zitkovich, Brianna and Yu, Tianhe and Xu, Sichun and Xu, Peng and Xiao, Ted and Xia, Fei and Wu, Jialin and Wohlhart, Paul and Welker, Stefan and Wahid, Ayzaan and others},
  booktitle={Conference on Robot Learning},
  pages={2165--2183},
  year={2023},
  organization={PMLR}
}

@inproceedings{octo,
  title={Octo: An open-source generalist robot policy},
  author={{Octo Model Team} and Ghosh, Dibya and Walke, Homer and Pertsch, Karl and Black, Kevin and Mees, Oier and Dasari, Sudeep and Hejna, Joey and Kreiman, Tobias and Xu, Charles and others},
  booktitle={Robotics: Science and Systems},
  year={2024}
}

@inproceedings{catchit,
  title={Catch it! learning to catch in flight with mobile dexterous hands},
  author={Zhang, Yuanhang and Liang, Tianhai and Chen, Zhenyang and Ze, Yanjie and Xu, Huazhe},
  booktitle={2025 IEEE International Conference on Robotics and Automation (ICRA)},
  pages={14385--14391},
  year={2025},
  organization={IEEE}
}

@article{zeng2020tossingbot,
  title={Tossingbot: Learning to throw arbitrary objects with residual physics},
  author={Zeng, Andy and Song, Shuran and Lee, Johnny and Rodriguez, Alberto and Funkhouser, Thomas},
  journal={IEEE Transactions on Robotics},
  volume={36},
  number={4},
  pages={1307--1319},
  year={2020},
  publisher={IEEE}
}

@inproceedings{flingbot,
  title={Flingbot: The unreasonable effectiveness of dynamic manipulation for cloth unfolding},
  author={Ha, Huy and Song, Shuran},
  booktitle={Conference on Robot Learning},
  pages={24--33},
  year={2022},
  organization={PMLR}
}

@inproceedings{dynamichandover,
  title={Dynamic handover: Throw and catch with bimanual hands},
  author={Huang, Binghao and Chen, Yuanpei and Wang, Tianyu and Qin, Yuzhe and Yang, Yaodong and Atanasov, Nikolay and Wang, Xiaolong},
  booktitle={Conference on Robot Learning},
  year={2023}
}

@inproceedings{tabletennis,
  title={Robotic table tennis: A case study into a high speed learning system},
  author={D'Ambrosio, David B and Abelian, Jonathan and Abeyruwan, Saminda and Ahn, Michael and Bewley, Alex and Boyd, Justin and Choromanski, Krzysztof and Cortes, Omar and Coumans, Erwin and Ding, Tianli and others},
  booktitle={Robotics: Science and Systems},
  year={2023}
}

@inproceedings{mstennis,
  title={Achieving human level competitive robot table tennis},
  author={DAmbrosio, David B and Abeyruwan, Saminda and Graesser, Laura and Iscen, Atil and Amor, Heni Ben and Bewley, Alex and Reed, Barney J and Reymann, Krista and Takayama, Leila and Tassa, Yuval and others},
  booktitle={2025 IEEE International Conference on Robotics and Automation (ICRA)},
  pages={74--82},
  year={2025},
  organization={IEEE}
}

@article{leave_no,
  title={Leave no observation behind: Real-time correction for vla action chunks},
  author={Sendai, Kohei and Alvarez, Maxime and Matsushima, Tatsuya and Matsuo, Yutaka and Iwasawa, Yusuke},
  journal={arXiv preprint arXiv:2509.23224},
  year={2025}
}

@article{smolvla,
  title={Smolvla: A vision-language-action model for affordable and efficient robotics},
  author={Shukor, Mustafa and Aubakirova, Dana and Capuano, Francesco and Kooijmans, Pepijn and Palma, Steven and Zouitine, Adil and Aractingi, Michel and Pascal, Caroline and Russi, Martino and Marafioti, Andres and others},
  journal={arXiv preprint arXiv:2506.01844},
  year={2025}
}

@article{dynamicvla,
  title={DynamicVLA: A Vision-Language-Action Model for Dynamic Object Manipulation},
  author={Xie, Haozhe and Wen, Beichen and Zheng, Jiarui and Chen, Zhaoxi and Hong, Fangzhou and Diao, Haiwen and Liu, Ziwei},
  journal={arXiv preprint arXiv:2601.22153},
  year={2026}
}

@article{f2fap,
  title={F2F-AP: Flow-to-Future Asynchronous Policy for Real-time Dynamic Manipulation},
  author={Wei, Haoyu and Xu, Xiuwei and Cheng, Ziyang and Yin, Hang and Ma, Angyuan and Yu, Bingyao and Zhou, Jie and Lu, Jiwen},
  journal={arXiv preprint arXiv:2604.02408},
  year={2026}
}

@inproceedings{videolavit,
  title={Video-lavit: Unified video-language pre-training with decoupled visual-motional tokenization},
  author={Jin, Yang and Sun, Zhicheng and Xu, Kun and Chen, Liwei and Jiang, Hao and Huang, Quzhe and Song, Chengru and Liu, Yuliang and Zhang, Di and Song, Yang and others},
  booktitle={International Conference on Machine Learning},
  year={2024}
}

@inproceedings{vidtwin,
  title={Vidtwin: Video vae with decoupled structure and dynamics},
  author={Wang, Yuchi and Guo, Junliang and Xie, Xinyi and He, Tianyu and Sun, Xu and Bian, Jiang},
  booktitle={Proceedings of the Computer Vision and Pattern Recognition Conference},
  pages={22922--22932},
  year={2025}
}

@inproceedings{reducio,
  title={Reducio! generating 1k video within 16 seconds using extremely compressed motion latents},
  author={Tian, Rui and Dai, Qi and Bao, Jianmin and Qiu, Kai and Yang, Yifan and Luo, Chong and Wu, Zuxuan and Jiang, Yu-Gang},
  booktitle={Proceedings of the IEEE/CVF International Conference on Computer Vision},
  pages={19237--19247},
  year={2025}
}

@inproceedings{sam2,
  title={Sam 2: Segment anything in images and videos},
  author={Ravi, Nikhila and Gabeur, Valentin and Hu, Yuan-Ting and Hu, Ronghang and Ryali, Chaitanya and Ma, Tengyu and Khedr, Haitham and R{\"a}dle, Roman and Rolland, Chloe and Gustafson, Laura and others},
  booktitle={International Conference on Learning Representations},
  year={2025}
}

@article{pi0,
  title={$\pi_0$: A Vision-Language-Action Flow Model for General Robot Control},
  author={Black, Kevin and Brown, Noah and Driess, Danny and Esmail, Adnan and Equi, Michael and Finn, Chelsea and Fusai, Niccolo and Groom, Lachy and Hausman, Karol and Ichter, Brian and others},
  journal={arXiv preprint arXiv:2410.24164},
  year={2024}
}

@inproceedings{brohan2022rt,
  title={RT-1: Robotics transformer for real-world control at scale},
  author={Brohan, Anthony and Brown, Noah and Carbajal, Justice and Chebotar, Yevgen and Dabis, Joseph and Finn, Chelsea and Gopalakrishnan, Keerthana and Hausman, Karol and Herzog, Alex and Hsu, Jasmine and others},
  booktitle={Robotics: Science and Systems},
  year={2023}
}

@inproceedings{driess2023palm,
  title={PaLM-E: An embodied multimodal language model},
  author={Driess, Danny and Xia, Fei and Sajjadi, Mehdi SM and Lynch, Corey and Chowdhery, Aakanksha and Ichter, Brian and Wahid, Ayzaan and Tompson, Jonathan and Vuong, Quan and Yu, Tianhe and others},
  booktitle={International Conference on Machine Learning},
  year={2023}
}

@article{chi2023diffusion,
  title={Diffusion policy: Visuomotor policy learning via action diffusion},
  author={Chi, Cheng and Xu, Zhenjia and Feng, Siyuan and Cousineau, Eric and Du, Yilun and Burchfiel, Benjamin and Tedrake, Russ and Song, Shuran},
  journal={The International Journal of Robotics Research},
  volume={44},
  number={10-11},
  pages={1684--1704},
  year={2025},
  publisher={Sage}
}

@inproceedings{zhao2023act,
  title={Learning fine-grained bimanual manipulation with low-cost hardware},
  author={Zhao, Tony Z and Kumar, Vikash and Levine, Sergey and Finn, Chelsea},
  booktitle={Robotics: Science and Systems},
  year={2023}
}

@misc{i2rt_yam_ultra,
  title        = {YAM Ultra 6-DoF Robotic Arm},
  author       = {{I2RT Robotics}},
  year         = {2024},
}

@inproceedings{zhao2024first,
  title={The first to know: How token distributions reveal hidden knowledge in large vision-language models?},
  author={Zhao, Qinyu and Xu, Ming and Gupta, Kartik and Asthana, Akshay and Zheng, Liang and Gould, Stephen},
  booktitle={European Conference on Computer Vision},
  pages={127--142},
  year={2024},
  organization={Springer}
}

@inproceedings{lu2025probing,
  title={Probing a vision-language-action model for symbolic states and integration into a cognitive architecture},
  author={Lu, Hong and Li, Hengxu and Shahani, Prithviraj Singh and Herbers, Stephanie and Scheutz, Matthias},
  booktitle={2025 IEEE International Conference on AI and Data Analytics (ICAD)},
  pages={1--8},
  year={2025},
  organization={IEEE}
}

@inproceedings{zhou2025vlm4d,
  title={Vlm4d: Towards spatiotemporal awareness in vision language models},
  author={Zhou, Shijie and Vilesov, Alexander and He, Xuehai and Wan, Ziyu and Zhang, Shuwang and Nagachandra, Aditya and Chang, Di and Chen, Dongdong and Wang, Xin Eric and Kadambi, Achuta},
  booktitle={Proceedings of the IEEE/CVF international conference on computer vision},
  pages={8600--8612},
  year={2025}
}

@inproceedings{li2024mvbench,
  title={Mvbench: A comprehensive multi-modal video understanding benchmark},
  author={Li, Kunchang and Wang, Yali and He, Yinan and Li, Yizhuo and Wang, Yi and Liu, Yi and Wang, Zun and Xu, Jilan and Chen, Guo and Luo, Ping and others},
  booktitle={Proceedings of the IEEE/CVF Conference on Computer Vision and Pattern Recognition},
  pages={22195--22206},
  year={2024}
}

@inproceedings{uvt,
  title={Unified Visuomotor Targets: Supervising VLAs Beyond Physical Actions},
  author={Feng, Zhenyang and Jain, Unnat},
  booktitle={IEEE/RSJ International Conference on Intelligent Robots and Systems (IROS)},
  year={2026}
}

@inproceedings{mvp,
  title={Real-World Robot Learning with Masked Visual Pre-training},
  author={Radosavovic, Ilija and Xiao, Tete and James, Stephen and Abbeel, Pieter and Malik, Jitendra and Darrell, Trevor},
  booktitle={Conference on Robot Learning},
  pages={416--426},
  year={2022},
  organization={PMLR}
}

@inproceedings{sensorimotor,
  title={Robot learning with sensorimotor pre-training},
  author={Radosavovic, Ilija and Shi, Baifeng and Fu, Letian and Goldberg, Ken and Darrell, Trevor and Malik, Jitendra},
  booktitle={Conference on Robot Learning},
  pages={683--693},
  year={2023},
  organization={PMLR}
}

@inproceedings{hpt,
  title={Scaling Proprioceptive-Visual Learning with Heterogeneous Pre-trained Transformers},
  author={Wang, Lirui and Chen, Xinlei and Zhao, Jialiang and He, Kaiming},
  booktitle={Advances in Neural Information Processing Systems},
  year={2024}
}

@inproceedings{r3m,
  title={R3M: A Universal Visual Representation for Robot Manipulation},
  author={Nair, Suraj and Rajeswaran, Aravind and Kumar, Vikash and Finn, Chelsea and Gupta, Abhinav},
  booktitle={Conference on Robot Learning},
  year={2022},
  organization={PMLR}
}

@inproceedings{roboaffordances,
  title={Affordances from Human Videos as a Versatile Representation for Robotics},
  author={Bahl, Shikhar and Mendonca, Russell and Chen, Lili and Jain, Unnat and Pathak, Deepak},
  booktitle={IEEE/CVF Conference on Computer Vision and Pattern Recognition (CVPR)},
  year={2023}
}

@inproceedings{data4robotics,
  title={An Unbiased Look at Datasets for Visuo-Motor Pre-Training},
  author={Dasari, Sudeep and Srirama, Mohan Kumar and Jain, Unnat and Gupta, Abhinav},
  booktitle={Conference on Robot Learning},
  year={2023}
}

@article{anchoralign,
  title={Generalizable VLA Finetuning via Representation Anchoring and Language-Action Alignment},
  author={Dalal, Dwip and Patel, Shivansh and Jain, Chahit and Kim, Jeonghwan and Mishra, Utkarsh and Baratian, Alex and Ha, Hyeonjeong and Ji, Heng and Lazebnik, Svetlana and Jain, Unnat},
  journal={arXiv preprint arXiv:2607.13429},
  year={2026}
}

\newpage
\appendix

\setcounter{figure}{0}
\setcounter{table}{0}
\renewcommand{\thefigure}{A\arabic{figure}}
\renewcommand{\thetable}{A\arabic{table}}

\section*{Appendix}
\addcontentsline{toc}{section}{Outline of the Appendix}

In this Appendix, we include extensions of analysis introduced in the main paper and additional technical details of our method, setup, and evaluation protocol. Qualitative rollouts, failure mode analysis, and more details are available on our project page: \url{https://tempo-robot.github.io/}.
\begin{itemize}
	\item \textbf{\cref{app:failure}, \emph{Additional Failure Mode Analysis}.} Supplements the trimming discussion in \cref{sec:exp-setup} and the success rates in \cref{tab:success_rates}. Explains why baselines collapse to $0\%$ on the original Flick Catch and Wine Pour, and reports \ours's performance on the untrimmed tasks (\cref{tab:state_alias}, \figref{fig:ambiguity}).
	\item \textbf{\cref{app:setup}, \emph{Experimental Setup Details}.} Supplements \cref{sec:exp-setup}. Lists the hardware platform, evaluation protocol, and training details for each method.
	\item \textbf{\cref{app:failure_breakdown}, \emph{Per-Task Failure Mode Breakdown}.} Supplements \cref{sec:quant} and the Bottle Handover breakdown in \figref{fig:failure_modes}. Extends the same per-rollout categorization to Flick Catch (\figref{fig:flick_failure}) and Drop Catch (\figref{fig:drop_failure}).
	\item \textbf{\cref{app:input_ablation}, \emph{Input Ablation}.} Supplements the single-signal ablations in \cref{sec:ablation}. Measures per-frame causal reliance on \TEMPOMOT\ and \TEMPOACT\ via input ablation MSE (\figref{fig:attention_mse}).
	\item \textbf{\cref{app:motion_encoder}, \emph{Motion Encoder Selection}.} Supplements the motion-encoder definition in \cref{sec:method} ($\phi$ in \TEMPOMOT). Justifies our choice of SAM~2.1-Tiny by linearly probing several candidate encoders for object velocity (\cref{fig:ball_velocity_probe}).
\end{itemize}

\section{Additional Failure Mode Analysis}
\label{app:failure}

\begin{figure}[h]
	\centering
	\includegraphics[width=\columnwidth]{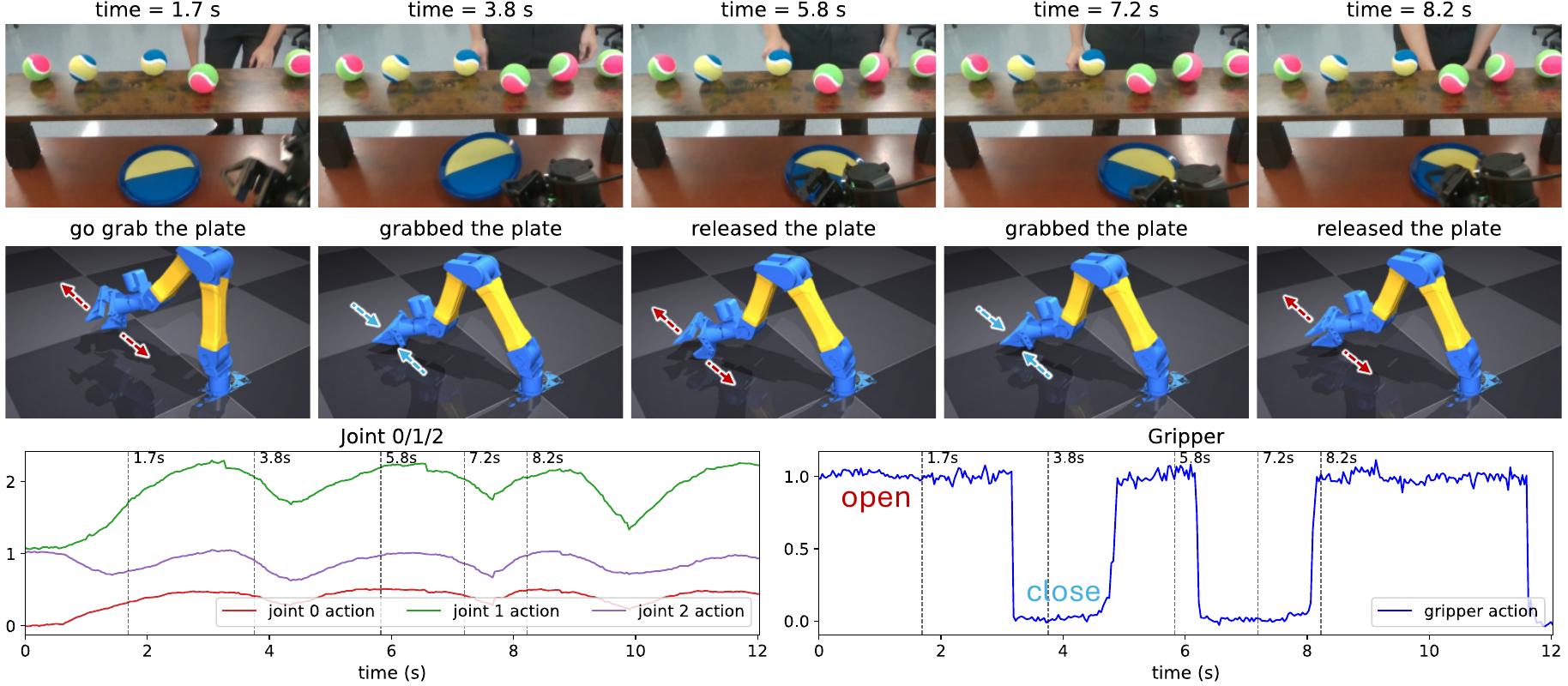}
	\caption{\textbf{VLASH rollout on the untrimmed Flick Catch task.} The task is to grip the velcro plate and follow the human hand. VLASH's gripper hovers near the plate without committing to grasp or retract, because a single-frame observation cannot differentiate reaching for the plate from releasing it. RTC exhibits the same failure (see project page).}
	\label{fig:ambiguity}
\end{figure}

\emph{Supplements the trimming discussion in \cref{sec:exp-setup} and \cref{tab:success_rates}; extends the state-aliasing evidence in \cref{sec:diagnosis}.} In our initial setup, every method was trained and evaluated on the complete, untrimmed versions of all four tasks. On Flick Catch and Wine Pour, both asynchronous baselines failed every rollout. The cause is state aliasing rather than reaction speed or accuracy: each episode contains a reach-out phase, in which the gripper approaches the velcro plate or the bottle, and a release-and-retract phase, in which the gripper recedes from the same object after grasping or pouring. The two phases produce visually near-identical observations yet demand opposite actions, and a single-frame policy has no cue to disambiguate them, so the gripper oscillates between extending toward and withdrawing from the object and never commits long enough to make progress (\figref{fig:ambiguity}; see qualitative rollouts on our project webpage).

Because both baselines collapse to $0\%$ on the untrimmed task, a direct comparison there is uninformative. To enable a non-trivial comparison that isolates motion ambiguity from state aliasing, we manually trim the release-and-retract segment from every episode of the Flick Catch and Wine Pour training and evaluation data; the success rates on this trimmed version are reported in \cref{tab:success_rates}. The trimmed task still requires motion-aware perception (target tracking and timing) but removes the state-aliased segment.

To verify that \ours\ resolves state aliasing rather than merely benefiting from this trimming, we additionally train and evaluate every method on the original, untrimmed episodes. \cref{tab:state_alias} (in \cref{sec:diagnosis}) reports the resulting success rates: RTC and VLASH again fail every rollout, while \ours\ succeeds at $68\%$ on Flick Catch and $97.6$\% on Wine Pour. We attribute the gain to \TEMPOACT: the proprioceptive history distinguishes task phases that the current frame cannot, letting the policy commit to the correct action rather than hedging between reach-out and retract.

\section{Experimental Setup Details}
\label{app:setup}

\emph{Supplements \cref{sec:exp-setup}.}

\subsection{Hardware Platform}
All experiments are conducted on a bimanual workstation consisting of two I2RT YAM Ultra robot arms~\cite{i2rt_yam_ultra}, each with 7 degrees of freedom. The two arms are mounted and clamped to the table edge with their bases separated by 24 inches (61 cm). Both arms are placed on a rigid table measuring 6 ft × 29.5 in × 28 in (width × depth × height; 183 × 75 × 71 cm). Each arm is equipped with a parallel-jaw gripper.

 Visual observations are provided by three cameras: one fixed third-person camera positioned between the two arms and two wrist-mounted cameras. Each camera streams 224p at 30 Hz. Policies are executed at 30 Hz, if possible.

\subsection{Evaluation Protocol}

For the Drop Catch task, a single ball is released above the workspace from a height of 13–18 in (33–46 cm), measured from the table surface to the release point. The ball is tennis-ball sized and covered in a velcro-compatible material so that it adheres to the velcro plate that the gripper holds.

For the Flick Catch task, 5–6 balls are arranged on a shelf measuring 31.5 × 7.9 × 8.5 in (80 × 20 × 21.6 cm; width × depth × height), positioned near the table edge opposite to where the arms are clamped. These balls are identical to the one used in the Drop Catch task.

To ensure a fair evaluation, the human operator releasing the ball performs each trial with eyes closed while wearing headphones, preventing them from anticipating the robot's position and biasing the release point toward it.

\subsection{Training Details}

Each policy is trained on eight NVIDIA RTX PRO 6000 Blackwell GPUs. Training a single policy takes approximately 2 hours. More specifically, with an effective batch size of $128$ the per-iteration time is $0.49$s for VLASH, $0.46$s for RTC, and $0.50$s for \ours. RTC does not have an openly available codebase, so we reimplement it faithfully from its paper.

\section{Per-Task Failure Mode Breakdown}
\label{app:failure_breakdown}

\emph{Supplements \cref{sec:quant} and \figref{fig:failure_modes}.} To complement the Bottle Handover analysis in \cref{sec:quant}~(\figref{fig:failure_modes}), we apply the same protocol to Flick Catch and Drop Catch: we visually examine all $50$ rollouts per method on the trimmed task and assign each to a success or to one failure category. The categories differ across tasks because each task has different failure modes.

\begin{figure}[h]
	\centering
	\includegraphics[width=\columnwidth]{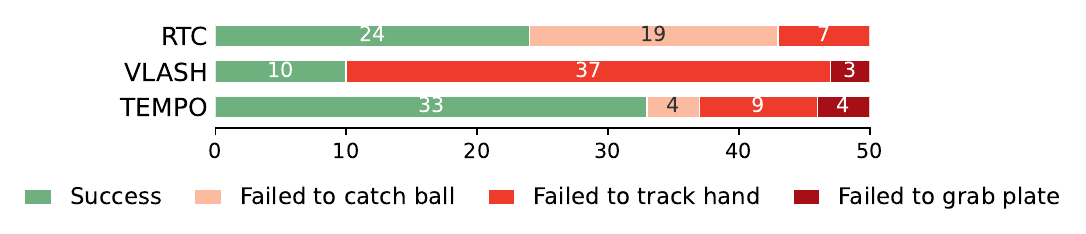}
	\caption{\textbf{Failure-mode breakdown on Flick Catch.} Each of the $50$ rollouts is decomposed into success plus three failure modes corresponding to the three task stages (grab plate, track hand, catch ball). VLASH's failures concentrate in hand-tracking and RTC struggles most at tracking the ball's trajectory (see project page); \ours~cuts both failure modes.}
	\label{fig:flick_failure}
\end{figure}

\paragraph{Flick Catch (\figref{fig:flick_failure}).}
We assign each failed rollout to one of three failure modes corresponding to the three task stages: grabbing the plate, tracking the hand, or catching the flicked ball. VLASH's failures concentrate almost entirely in hand-tracking: in $37$ of $50$ trials the policy fails to follow the human hand and therefore never reaches the catching stage. RTC reacts to the hand but loses the flicked ball, with ball-catch failures accounting for $19$ trials, the dominant mode for RTC. \ours\ cuts VLASH's hand-tracking failures from $37$ to $9$ and RTC's ball-catch failures from $19$ to $4$; the residual failures are spread across all three stages rather than concentrated in any single one.

\begin{figure}[h]
	\centering
	\includegraphics[width=\columnwidth]{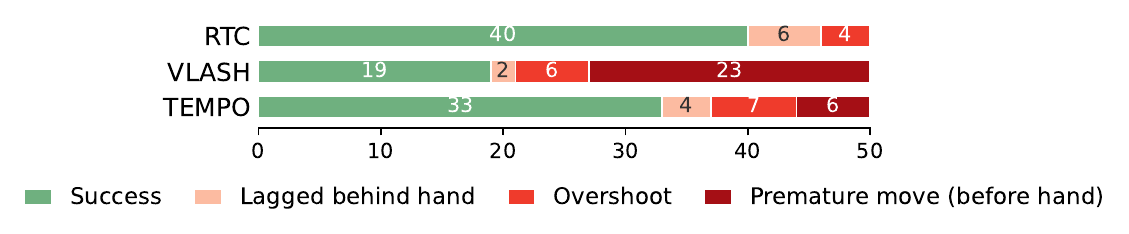}
	\caption{\textbf{Failure-mode breakdown on Drop Catch.} RTC scores highest as Drop Catch is dominated by reactive timing rather than motion or state ambiguity. VLASH's dominant failure is premature movement before the human's hand begins to move; \ours\ cuts this failure by nearly $4\times$.}
	\label{fig:drop_failure}
\end{figure}

\paragraph{Drop Catch (\figref{fig:drop_failure}).}
Drop Catch is the timing-dominated task on which RTC scores highest~(\cref{tab:success_rates}); its only failures are a few overshoots ($4$) and lags ($6$). VLASH, despite also being asynchronous, exhibits a markedly different profile: it \emph{moves before the human's hand} in $23$ of $50$ trials, a near-majority failure that pulls its success rate down to $19$. \ours\ cuts this premature-move failure by nearly $4\times$ (from $23$ to $6$), with only a small change in overshoots ($6\to 7$). The premature-move reduction is consistent with the role of \TEMPOACT: the proprioceptive history records what the policy has already begun, so it does not re-initiate a move on every frame.

\begin{figure}[h]
	\centering    \includegraphics[width=0.9\columnwidth]{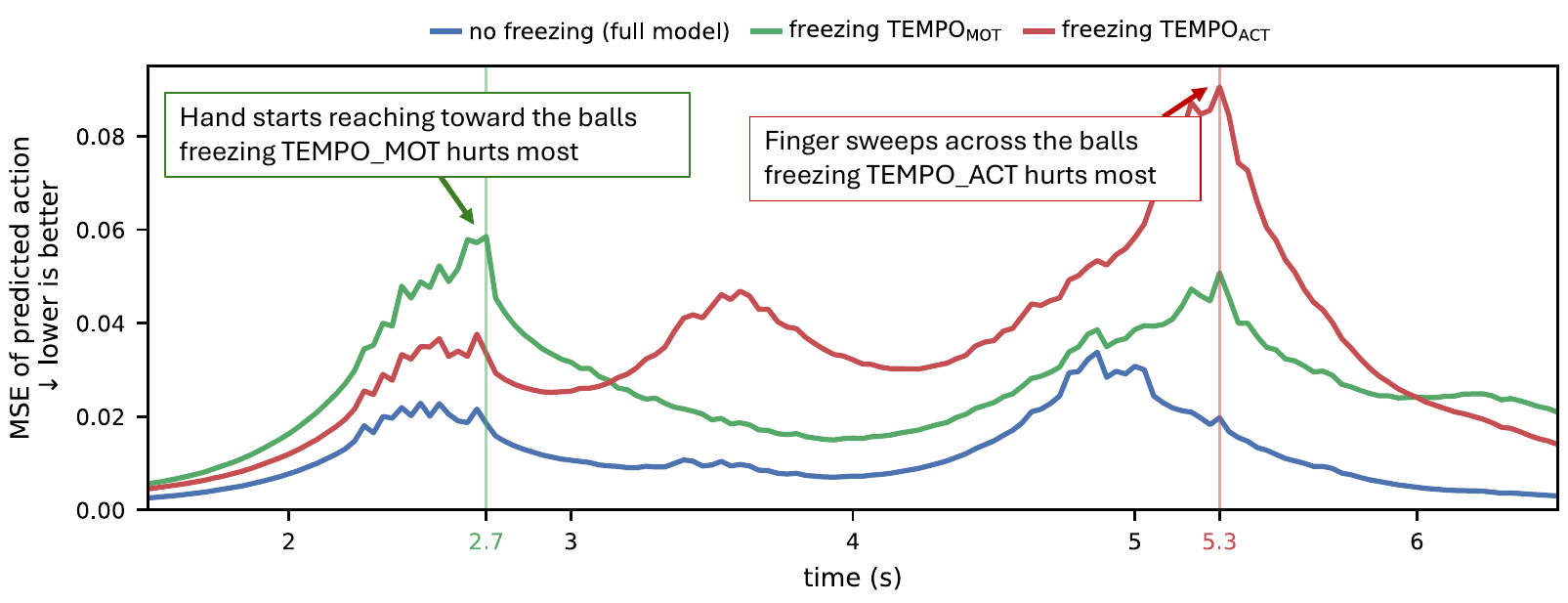}
	\vspace{-0.6em}
    \caption{\textbf{Both \TEMPOMOT\ and \TEMPOACT\ are necessary.} Per-frame action MSE for the full \ours\ model (blue) vs. the two single-signal ablations, evaluated on the Flick Catch task. Freezing \TEMPOMOT\ to its first-frame value (green) hurts most at $2.7$\,s, the beginning of the task where human hand starts to move. Freezing \TEMPOACT\ (red) hurts most at $5.3$\,s, the middle of the task where the finger sweeps horizontally across the balls before picking one and flicking it. Thus, each signal plays an important role at a different moment in the task.}
	\label{fig:attention_mse}
\end{figure}

\section{Input Ablation}
\label{app:input_ablation}

\emph{Supplements the modality-ablation analysis in \cref{sec:ablation}.}
To test whether \TEMPOMOT\ and \TEMPOACT\ are both necessary and contribute meaningfully to task performance, we run a per-frame ablation on a representative episode of the Flick Catch task (\figref{fig:attention_mse}). For each frame we ``remove'' each input modality by freezing \TEMPOMOT\ or \TEMPOACT\ to its first-frame value and measure the resulting predicted-action MSE; curves are EMA-smoothed for readability. Freezing either modality substantially degrades the policy's action prediction at different points in the episode. Freezing \TEMPOMOT\ hurts most at $2.7$\,s, the beginning of the task where large motions occur: MSE rises from $0.019$ to $0.059$ ($3.1\times$), while freezing \TEMPOACT\ costs only $0.033$. Freezing \TEMPOACT\ hurts most at $5.3$\,s, the middle of the task where the finger sweeps horizontally across the balls before picking one and flicking it, and there the pattern reverses: freezing \TEMPOACT\ hurts more than \TEMPOMOT\ ($0.091$ vs.\ $0.051$; full $0.020$). Each signal helps a different phase of the task.

\section{Motion Encoder Selection}
\label{app:motion_encoder}

\begin{figure}[h]
\centering

\begin{minipage}[b]{0.235\columnwidth}
	\centering
	\includegraphics[width=\linewidth]{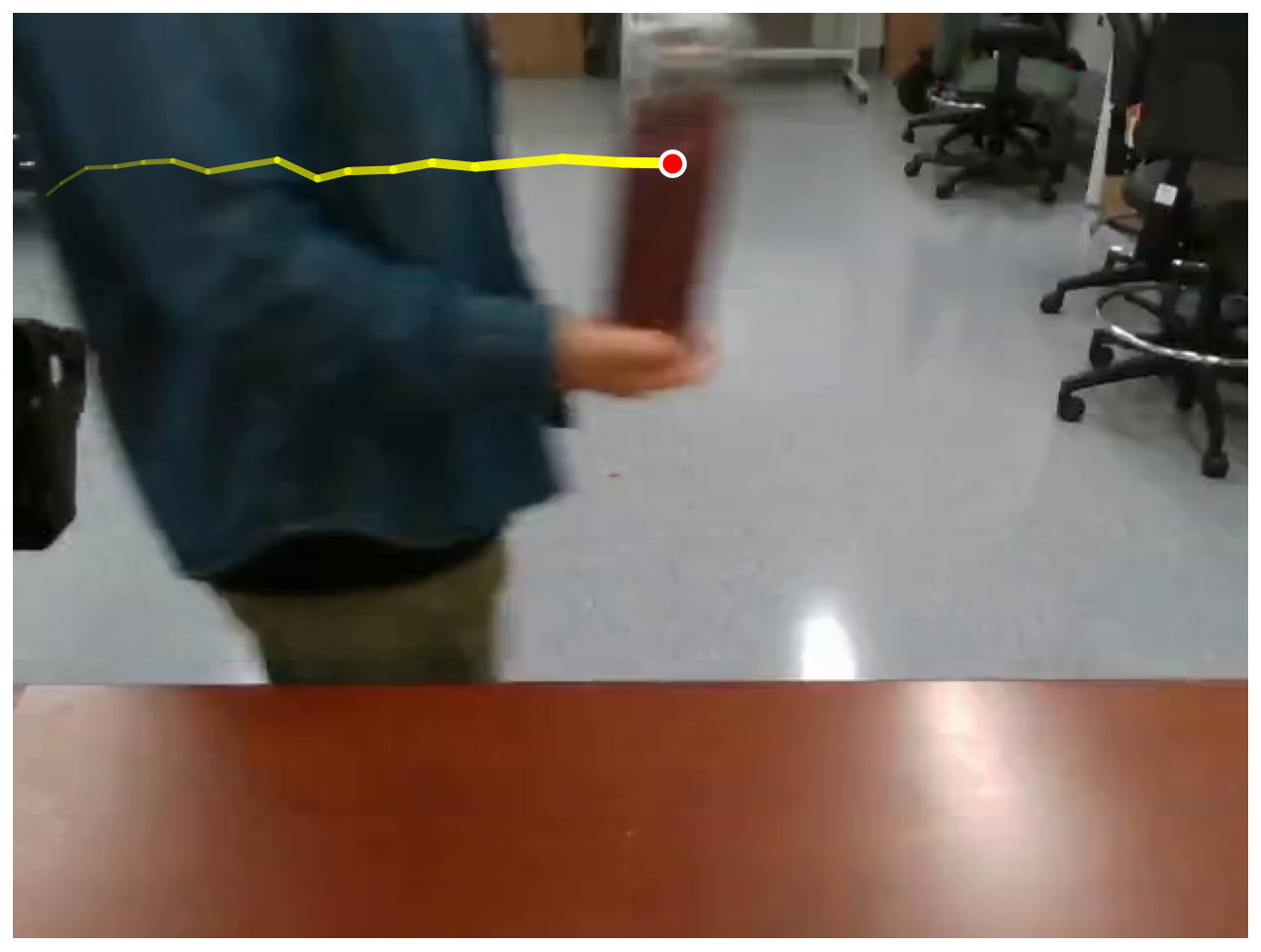}\\
	{\small Bottle Handover}
\end{minipage}\hfill
\begin{minipage}[b]{0.235\columnwidth}
	\centering
	\includegraphics[width=\linewidth]{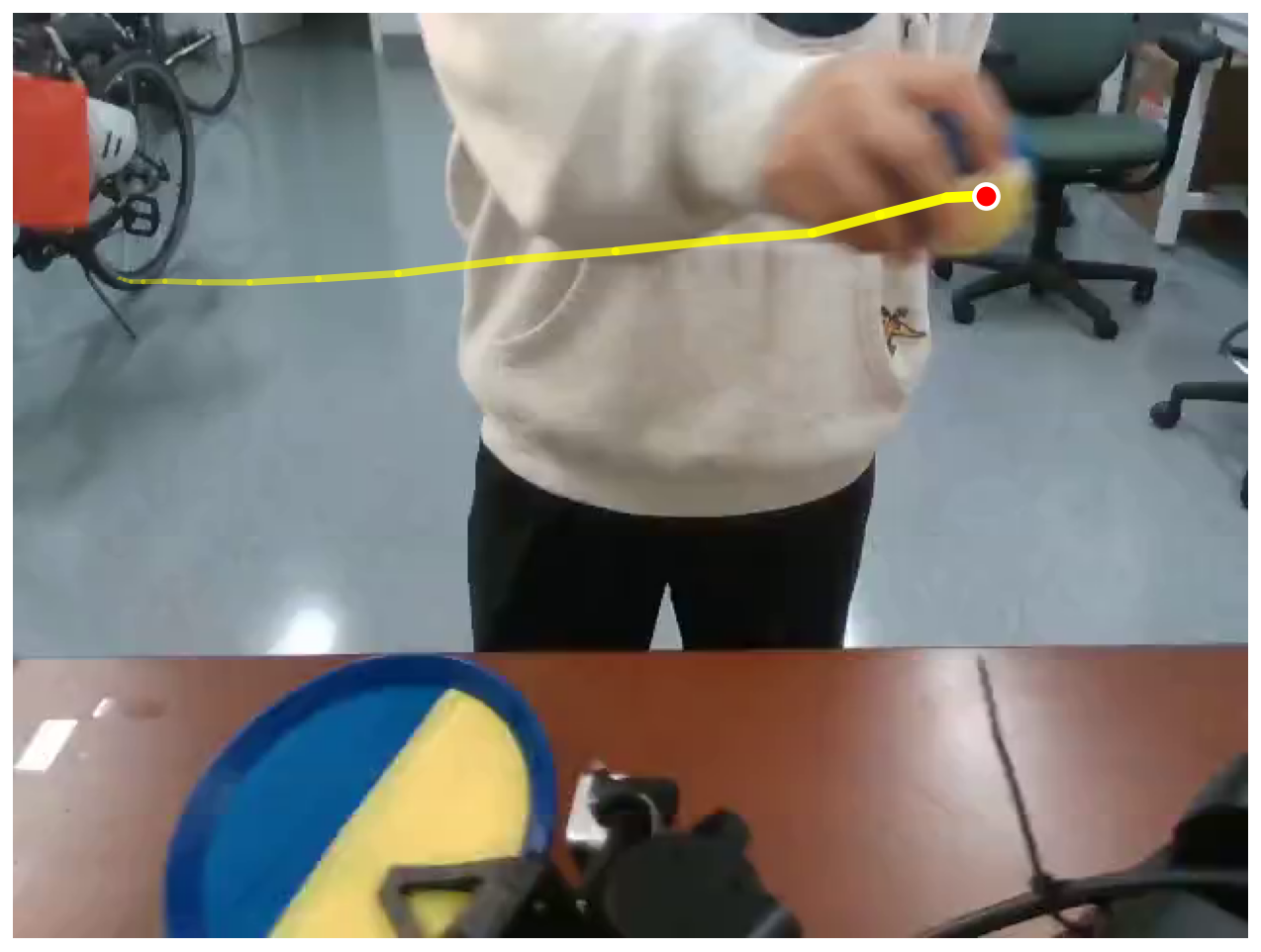}\\
	{\small Drop Catch}
\end{minipage}\hfill
\begin{minipage}[b]{0.235\columnwidth}
	\centering
	\includegraphics[width=\linewidth]{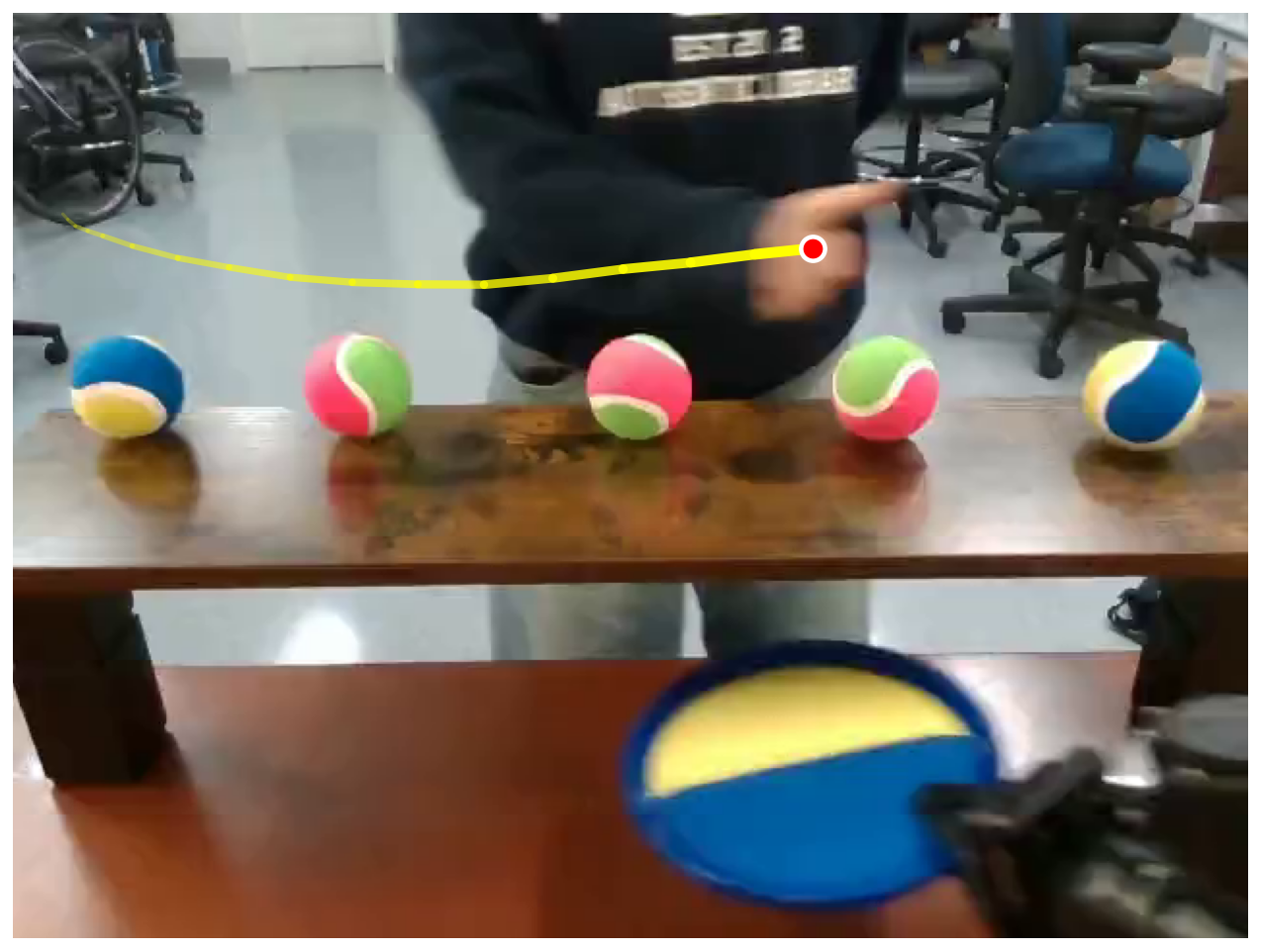}\\
	{\small Flick Catch}
\end{minipage}\hfill
\begin{minipage}[b]{0.235\columnwidth}
	\centering
	\includegraphics[width=\linewidth]{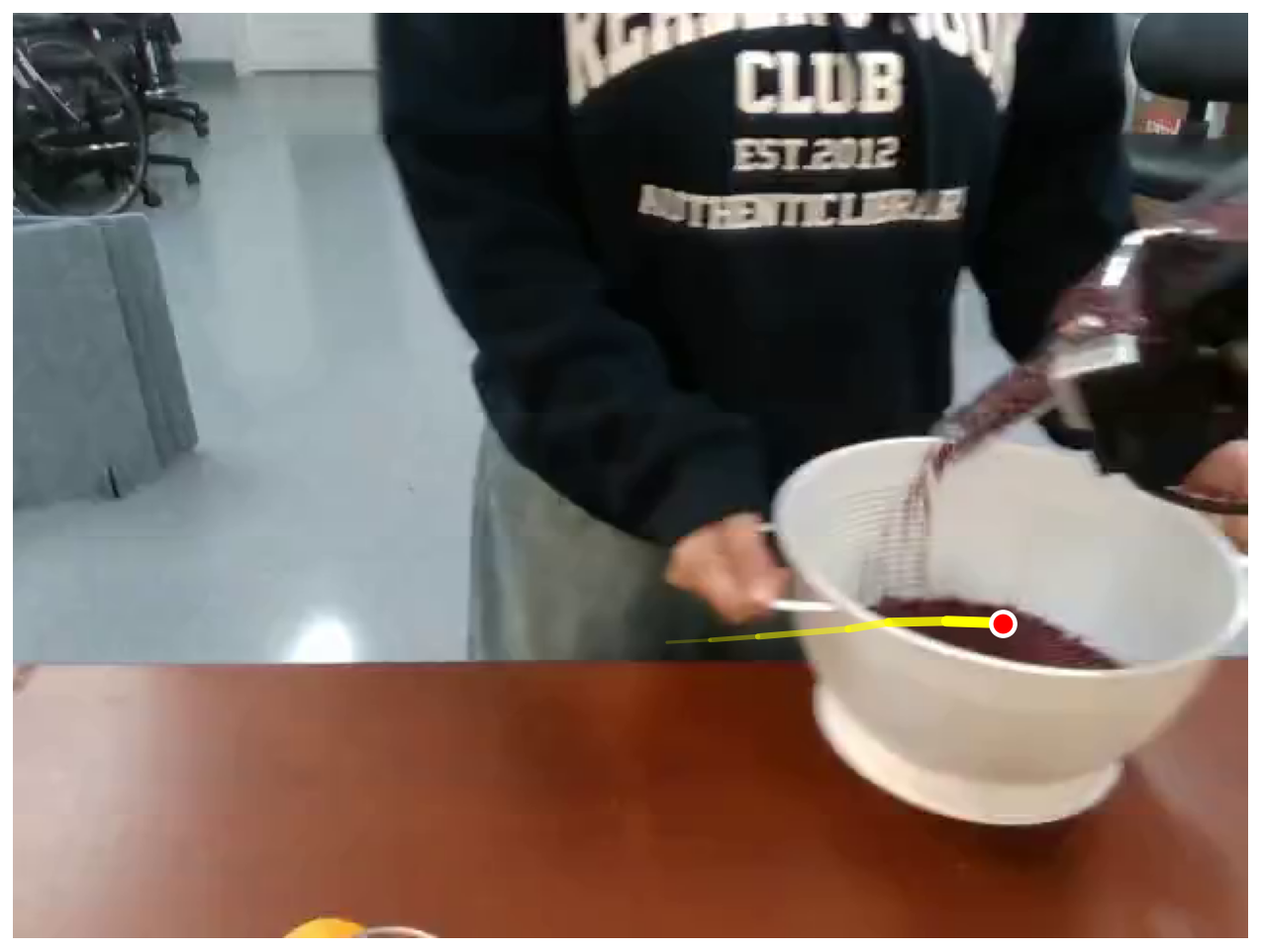}\\
	{\small Wine Pour}
\end{minipage}

\vspace{8pt}

\small
\begin{tabular}{lccccc}
\toprule
\textbf{Initialization for \TEMPOMOT} & \textbf{Bottle Handover} & \textbf{Drop Catch} & \textbf{Flick Catch} & \textbf{Wine Pour} & \textbf{Average} \\
\midrule
Object Position                          & 0.24  & 0.03 & 0.03 & 0.02 & 0.08 \\
VideoLaVIT~\cite{videolavit}             & 0.45  & 0.25 & 0.28 & 0.36 & 0.34 \\
Reducio~\cite{reducio}                   & 0.50  & 0.36 & 0.38 & 0.32 & 0.39 \\
SAM~2~\cite{sam2}                        & 0.58  & 0.37 & 0.50 & 0.39 & 0.46 \\
VidTwin~\cite{vidtwin}                   & 0.59  & 0.44 & 0.43 & 0.45 & 0.48 \\
\bottomrule
\end{tabular}

\caption{\textbf{Linear probe of candidate initializations for \TEMPOMOT.} \emph{Top:} one example frame per task showing the object of interest (red dot) and its recent trajectory (yellow line); the pixel-space velocity of this object is the regression target the linear probe must predict. \emph{Bottom:} per-task velocity $R^2$ for each candidate initialization, along with the average. Higher is better; the raw object-position baseline ($R^2=0.08$ on average) cannot recover velocity from a single position, while every learned motion encoder does so substantially better.}
\label{fig:ball_velocity_probe}
\end{figure}

\emph{Supplements the motion-encoder definition in \cref{sec:method} (the $\phi$ in \TEMPOMOT).}
\ours\ is agnostic to the choice of motion encoder; here we justify ours. Reusing the per-frame object-velocity annotation from our motion probe (\secref{sec:analysis}; the per-task probing target is visualized at the top of \figref{fig:ball_velocity_probe}), we fit a \emph{linear} probe from each candidate motion latent to the object's velocity: if the latent captures the object's motion, a linear readout should recover the velocity with high $R^2$. Every learned encoder decodes velocity far above the raw object position (avg.\ $R^2 = 0.08$), confirming that object motion is genuinely present in the latent and that a single position, like a single frame, does not by itself encode meaningful information to infer object motion. SAM~2 and VidTwin score highest; we adopt SAM~2.1-Tiny because it pairs near-top motion correlation with the streaming speed required for real-time deployment (\cref{tab:latency_comparison}).


\end{document}